\pdfoutput=1

\documentclass[11pt]{article}

\usepackage[]{EMNLP2022}

\usepackage{times}
\usepackage{latexsym}
\usepackage{graphicx}
\usepackage{booktabs, multirow} 
\usepackage{amsmath}
\usepackage{xspace,mfirstuc,tabulary}
\usepackage[inline]{enumitem}

\usepackage[T1]{fontenc}

\usepackage[utf8]{inputenc}

\usepackage{microtype}

\usepackage{inconsolata}

\newcommand{\ignore}[1]{}
\newenvironment{itemizesquish}{\begin{list}{\labelitemi}{\setlength{\itemsep}{0em}\setlength{\labelwidth}{0.7em}\setlength{\leftmargin}{\labelwidth}\addtolength{\leftmargin}{\labelsep}}}{\end{list}}

\newcommand{\bto}[1]{}
\newcommand{\forjg}[1]{}
\newcommand{\ag}[1]{}
\newcommand{\scomment}[1]{}
\newcommand{\jhg}[1]{}
\newcommand{\mk}[1]{}

\newcommand{\factbank}{{F}act{B}ank\xspace}
\newcommand{\bert}{{BERT}\xspace}
\newcommand{\polibelief}{{P}oli{B}elief\xspace}
\newcommand{\timebank}{{T}ime{B}ank\xspace}
\newcommand{\defacto}{{D}e{F}acto\xspace}
\newcommand{\nonepistemic}{\emph{{N}on-{E}pistemic}\xspace}

\title{Examining Political Rhetoric with Epistemic Stance Detection}

\author{Ankita Gupta$^{\diamondsuit}$ \quad Su Lin Blodgett$^{\heartsuit}$ \quad Justin H Gross$^\diamondsuit$ \quad Brendan O'Connor$^\diamondsuit$\\\\
$^\diamondsuit$University of Massachusetts Amherst, $^\heartsuit$Microsoft Research \\ \texttt{\{ankitagupta,jhgross,brenocon\}@umass.edu}\\ \texttt{sublodge@microsoft.com} }

\begin{document}
\maketitle

\begin{abstract}

Participants in political discourse employ rhetorical strategies---such as hedging, attributions, or denials---to display varying degrees of belief commitments to claims proposed by themselves or others. 
Traditionally, political scientists have studied these epistemic phenomena through labor-intensive manual content analysis.
We propose to help automate such work through epistemic stance prediction, drawn from research in computational semantics, 
to distinguish at the clausal level what is asserted, denied, or only ambivalently suggested 
by the author or other mentioned entities (\emph{belief holders}).
We first develop a simple RoBERTa-based model for multi-source stance predictions that outperforms more complex state-of-the-art modeling.
Then we demonstrate its novel application to political science by conducting a large-scale analysis of the Mass Market Manifestos corpus of U.S.\ political opinion books,
where we characterize trends in cited belief holders---respected allies and opposed bogeymen---across U.S.\ political ideologies.



\end{abstract}

\section{Introduction}
\label{sec:intro}

\begin{figure*}[t]
    \centering
   \includegraphics[width=120mm, height=40mm, scale=0.8,clip,trim=0 0 0 0mm]{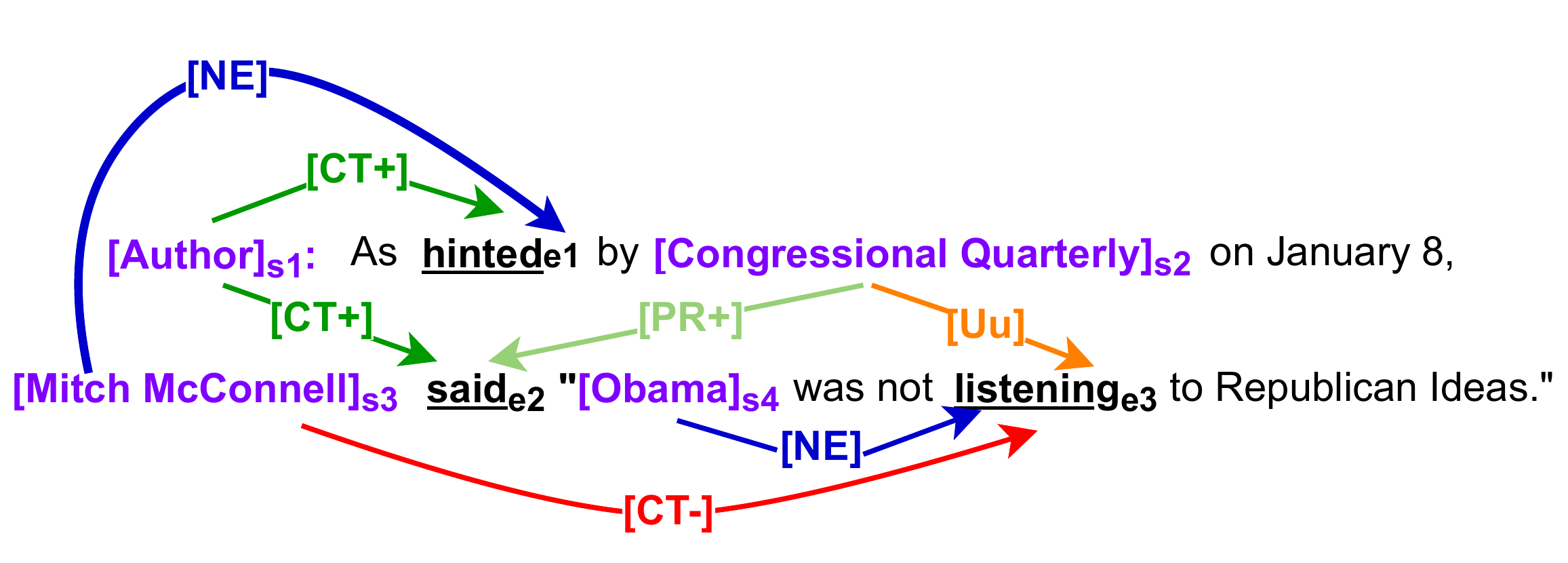}
  \caption{Illustrative example, simplified and adapted from a sentence in the Mass Market Manifestos corpus. 
  There are four sources (s1--s4) and three events (e1--e3) with $4\times3=12$ labels between them;
  all epistemic stances are shown, but most non-epistemic (NE) labels are hidden for clarity.
  \S\ref{sec:intro} and \S\ref{sec:epistemic_stance_framework} describe the labels.
  \label{fig:motivating_example}}
\end{figure*}

Political argumentation is rich with assertions, hypotheticals and disputes over opponent's claims. While making these arguments, political actors often employ several rhetorical strategies to display varying degrees of commitments to their claims. For instance, political scientists have studied the \textit{footing-shift} strategy, where actors convey their own beliefs while claiming that they belong to someone else~\cite{goffman1981forms, clayman1992footing}. 
Sometimes they may attribute their beliefs to a majority of the population via \textit{argument from popular opinion}\ \cite{walton2008argumentation}. 
Actors can also resort to \textit{hedging}, stating their own beliefs, but qualified with a partial degree of certainty~\cite{fraser2010hedging, lakoff1975hedges, hyland1996writing}
or express simple \textit{political disagreements}, 
contradicting claims made by their opponents~\cite{jang2009diverse, klofstad2013disagreeing, frances2014disagreement, christensen2009disagreement}.

Traditionally, political scientists and other scholars have manually analyzed
the impact of such strategies and argumentation on audience perception~\cite{clayman1992footing, fraser2010hedging}.
Recent advances in natural language processing (NLP) and digital repositories of political texts have enabled researchers to conduct large-scale analyses of political arguments using methods such as subjectivity analysis~\cite{liu2012sentiment, pang2009opinion}, argument mining~\cite{DBLP:conf/aaai/TrautmannDSSG20, toulmin-1958-the, walton1996argumentation}, and opinion mining~\cite{wiebe2005annotating, bethard2004automatic, kim-hovy-2004-determining, choi-2005-identifying}. While these approaches primarily concern argument structure and normative attitudes, we propose a complementary approach to analyze sources' \emph{epistemic} attitudes towards assertions~\cite{langacker-2009-investigations, anderson-1986-evidentials, arrese-2009-effective}---what they believe to be true and the extent to which they commit to these beliefs. 

Consider an example shown in Figure~\ref{fig:motivating_example}, where the author of the text (s1) quotes a speculation from the Congressional Quarterly (s2) about what Mitch McConnell (s3) said concerning Obama (s4).
In this example, while the author of the text believes that the Congressional Quarterly hinted something about McConnell (thus, exhibiting a \emph{certainly positive} (\textit{CT+}) stance towards the event (e1),
she remains \emph{uncommitted} (\textit{Uu}) 
about the quoted event (e3) that McConnell describes (edge omitted for visual clarity).
Of course, this event is asserted as \emph{certainly negative} (\textit{CT-})
by McConnell, the speaker of the quote.
The Congressional Quarterly suggests that Mitch McConnell made a statement (a \emph{probably positive} (\textit{PR+}) stance towards e2) while remaining \emph{uncommitted} towards what he said. 
Finally, \emph{Obama}'s own beliefs about whether he paid attention to Republican ideas are not expressed in this sentence; thus, s4 (Obama) has a \emph{non-epistemic} label toward the listening event (e3).


To address this challenging problem of epistemological analysis, researchers within the NLP community have created several datasets and models in various domains~\cite{minard2016meantime, Rambow2016BeSt, rudinger2018neural, lee2015event, stanovsky2017integrating, white2016universal, de-marneffe-etal-2012-happen}, often motivated directly by the interesting challenges of these linguistic semantic phenomena.
However, there is a great potential to use an epistemic stance framework to analyze social relations~\cite{soni2014modeling, prabhakaran2015new, swamy2017have}, motivating us to further advance this framework to support analysis of common rhetorical strategies and argumentation styles used in political discourse.

In this paper, we seek to further how \textit{epistemic stance} analysis can help computationally investigate the use of \textit{rhetorical strategies} employed in political discourse. 
In particular, we use the theory, structure and annotations of \factbank \cite{sauri2009factbank}, an expert-annotated corpus drawn
from English news articles,
which distinguishes different types of epistemic stances expressed in text.
\factbank features annotations not just for the author, but also other sources (entities) mentioned in the text. Such multi-source annotations allow us to disambiguate the author's own beliefs from the beliefs they attribute to others.

Our main contributions in this work are:
\begin{itemize}

\item We conduct a literature review connecting ideas related to epistemic stance as studied across several disconnected scholarly areas of linguistics, NLP, and political science (\S\ref{sec:related_work}).

 \item We develop a fine-tuned RoBERTa model~\cite{liu2019roberta} for multi-source epistemic stance prediction (\S\ref{sec:model}), whose simplicity makes it accessible to social scientist users,\footnote{All resources accompanying this project are added to our project page: \url{https://github.com/slanglab/ExPRES}} while performing at par with a more complex state-of-the-art model~\cite{qian-2018-gan}.

 
 
 
 \item We use our model to identify the most frequent \emph{belief holders} which are epistemic sources whose views or statements are expressed by the author. Identifying belief holders is
 an essential first step in analyzing rhetorical strategies and arguments.
We conduct this study on the Mass-Market Manifestos (MMM) Corpus,
a collection of 370 contemporary English-language political books authored by an ideologically diverse group of U.S. political opinion leaders. We compare results to traditional named entity recognition. Finally, we analyze differences in what belief holders tend to be cited by left-wing versus right-wing authors, revealing interesting avenues for future work in the study of U.S.\ political opinion
(\S\ref{sec:case_study}).

\item  In the appendix, we additionally validate our model by replicating an existing manual case study comparing the commitment levels of different political leaders\ (\S\ref{sec:appendix_hedging}, \citealp{jalilifar2011power}), and give further analysis of the model's behavior with negative polarity items and different types of belief holders (\S\ref{sec:appendix_bh}).
\end{itemize}

\section{Epistemic Stance from Different Perspectives}
\label{sec:related_work}
The notion of epistemic stances has been studied under several scholarly areas, including linguistics, political science and NLP. In this section, we discuss various notions of epistemic stances and how they have been utilized in these different areas.

\subsection{Epistemic Stance in Linguistics}
A speaker's \emph{epistemic stance} is their positioning about their knowledge of, or veracity of, communicated events and assertions~\cite{biber-1989-styles, palmer-2001-mood, langacker-2009-investigations}. Epistemic stance relates to the concept of \textit{modality}, which deals with the degree of certainty of situations in the world, and has been extensively studied under linguistics~\cite{kiefer1987defining, palmer-2001-mood, lyons1977semantics, chafe-1986-evidentiality} and logic~\cite{horn1972semantic, hintikka1962knowledge, hoek1990systems, holliday2018epistemic}. 
From a cognitivist perspective, epistemic stance concerns the pragmatic relation between
speakers and their knowledge regarding assertions~\cite{biber-1989-styles, mushin2001evidentiality, martin2005appraisal}.

\subsection{Epistemic Stance in Political Science}
The use of epistemic stances is widespread in political communication and persuasive language, to describe assertions when attempting to influence the reader's view~\cite{chilton-2004-analysing, arrese-2009-effective}. For instance, \citet{chilton-2004-analysing} studies use of epistemic stances by speakers/writers for legitimisation and coercion; \citet{arrese-2009-effective} examines epistemic stances taken by speakers to reveal their ideologies. In these studies, a speaker's communicated stance may follow what they believe due to their experiences, inferences, and mental state \cite{anderson-1986-evidentials}. From a psychological perspective, ~\citet{shaffer1981balance} employs balance theory~\cite{heider1946attitudes}---the cognitive effect of knowing an entity's stance towards an issue---in explaining public perceptions of presidential candidates' issue positions. 


\subsection{Epistemic Stance in NLP}
\label{sec:epistemic_stance_in_nlp}
In the NLP literature, epistemic stances---typically of authors, and sometimes of mentioned textual entities---have been studied 
under the related concepts of
\emph{factuality} \cite{sauri2012you, rudinger-etal-2018-neural-davidsonian, lee2015event, stanovsky2017integrating, minard2016meantime,soni2014modeling}
and 
\emph{belief commitments} \citep{prabhakaran2015new,diab2009committed}. \citet{de-marneffe-etal-2012-happen} prefers the term \emph{veridicality} to study the reader's, not author's, perspective.

We use the term \emph{epistemic stance} to avoid confusion with 
at least two more recent subliteratures that use \emph{factuality} differently from the above.
In misinformation detection, factuality refers to a proposition's objective truth \cite{rashkin-etal-2017-truth, mihaylova2018fact, thorne2018fever, vlachos2014fact}.
By contrast, we follow the epistemic stance approach in not assuming any objective reality---we simply model whatever subjective reality that agents assert.
Furthermore, text generation work has studied whether text summaries
conform to a source text's asserted propositions---termed the factuality or ``factual correctness'' of a summary \cite{maynez-etal-2020-faithfulness, wiseman-etal-2017-challenges, kryscinski-etal-2019-neural, dhingra2019handling}.

Several researchers in NLP have explored interesting social science applications in multiple settings such as organizational interactions~\cite{prabhakaran2010automatic}, Supreme Court hearings~\cite{danescu2012echoes}, discussion~\cite{bracewell2012motif, swayamdipta2012pursuit} and online forums~\cite{biran2012detecting, rosenthal2014detecting}. In particular, \citet{prabhakaran2010automatic} use epistemic stances to analyse power relations in organizational interactions. These studies demonstrate the potential of using epistemic stance analysis for social science applications. Motivated by these advances, we use epistemic stance framework to analyze political rhetoric, a genre that has not been explored earlier.

\begin{table}[]
\centering \tiny
\begin{tabular}{p{0.65cm}llll}
\hline \addlinespace[0.05cm]
\textbf{Type} &
  \textbf{Dataset} &
  \textbf{Perspective} &
  \textbf{Genre} &
  \textbf{Label} \\ \addlinespace[0.05cm]  \hline \addlinespace[0.05cm]
\multirow{5}{*}{Factuality} &
  \begin{tabular}[c]{@{}l@{}}\factbank \\ \cite{sauri2012you}\end{tabular} &
  Multi &
  News &
  Disc (8) \\ \addlinespace[0.1cm]
 &
  \citealp{stanovsky2017integrating} &
  Author &
  News &
  Cont [-3, 3] \\ \addlinespace[0.1cm]
 &
  \begin{tabular}[c]{@{}l@{}}MEANTIME \\ \cite{minard2016meantime}\end{tabular} &
  Multi &
   \begin{tabular}[c]{@{}l@{}}  News \\ (Italian)  \end{tabular} &
  Disc (3) \\ \addlinespace[0.1cm]
 &
  \citealp{lee2015event} &
  Author &
  News &
  Cont [-3, 3] \\ \addlinespace[0.1cm]
 &
  \begin{tabular}[c]{@{}l@{}}UDS-IH2 \\ \cite{rudinger2018neural}\end{tabular} &
  Author &
  Open &
  \begin{tabular}[c]{@{}l@{}}Disc (2) \&\\ Conf [0,4]\end{tabular} \\ \addlinespace[0.1cm]
  &
  \citealp{yao-2021-mds} &
  Multi &
  News &
  Disc (6) \\ \addlinespace[0.1cm]
   &
  \citealp{vigus-2019-dependency} &
  Multi &
  Open &
  Disc (6) \\ \addlinespace[0.1cm]
  \hline \addlinespace[0.1cm]
\begin{tabular}[c]{@{}l@{}}Indirect \\ Reporting\end{tabular} &
  \citealp{soni2014modeling} &
  Reader &
  Twitter &
  Likert (5) \\ \addlinespace[0.1cm] \hline \addlinespace[0.1cm]
\begin{tabular}[c]{@{}l@{}}Pragmatic \\ Veridicality\end{tabular} &
  \begin{tabular}[c]{@{}l@{}}PragBank \\ \cite{de-marneffe-etal-2012-happen}\end{tabular} &
  Reader &
  News &
  Disc (7) \\ \addlinespace[0.1cm] \hline \addlinespace[0.1cm]
\multirow{2}{*}{Beliefs} &
  \citealp{diab2009committed} &
  Author &
  Open &
  Disc (3) \\
 &
  \citealp{prabhakaran2015new} &
  Author &
  Forums &
  Disc (4) \\ \addlinespace[0.1cm] \hline
\end{tabular}
\caption{Summary of epistemic stance annotated datasets. \emph{Perspective}: which sources are considered for annotation? Stance \emph{Label} may be discrete with the given number of categories (where many or all are ordered), or continuous with a bounded range.\textsuperscript{\ref{fn:datasets}}
All datasets except MEANTIME consist of English text. 
\label{tab:datasets}
}
\vspace{-0.5cm}
\end{table}

\paragraph{Existing Datasets} Several existing datasets \cite{rudinger2018neural, lee2015event, prabhakaran2015new, diab2009committed, stanovsky2017integrating} have successfully driven the progress of epistemic stance analysis in NLP, but have largely focused on author-only analysis.~\citet{soni2014modeling} and \citet{de-marneffe-etal-2012-happen}~examine epistemic stances from the reader's (not author's) perspective.~Table~\ref{tab:datasets}~summarizes these datasets.\footnote{\label{fn:datasets}UDS-IH2 collects a binary category and a confidence score. \citet{yao-2021-mds} and \citet{vigus-2019-dependency} extend multisource annotations as dependency graphs with additional edge types.} 

Political discourse is a particularly interesting because the multiple sources discussed can have diverse stances towards the same event. Among all existing datasets, \factbank~\cite{sauri2012you} and MEANTIME~\cite{minard2016meantime} explore multi-source analysis in the news domain. While MEANTIME has helped advance epistemic stance analysis in Italian, \factbank---built on English news text---is closest to our goal.



\paragraph{Existing Models}
\label{sec:existing_models}
Several computational models have been developed for epistemic stance prediction as explicated in Table~\ref{tab:factuality_models}. Early models proposed deterministic algorithms based on hand-engineered implicative signatures for predicate lexicons \cite{lotan2013truthteller, nairn2006computing, sauri2012you}. A number of systems used lexico-syntactic features with supervised machine learning models, such as SVMs or CRFs \cite{diab2009committed, prabhakaran2010automatic, lee2015event, sauri2012you, stanovsky2017integrating}.
Lately, there has been a growing interest in using neural models for epistemic stance prediction \cite{rudinger2018neural, pouran-ben-veyseh-etal-2019-graph},
though sometimes with complex, task-specific network architectures (e.g.\ GANs; \citet{qian-2018-gan}),
which raise questions about generalization and replicability for practical use by experts from other fields.
Recently, \citet{jiang2021thinks} explore fine-tuning pre-trained language models (LM), such as \bert, for author-only epistemic stance prediction by adding a simple task-specific layer. We take this more robust approach, extending it to multiple sources.

\begin{table}[]
\centering \tiny
\begin{tabular}{llll}
\hline \addlinespace[0.05cm]
\textbf{Algorithm} &
  \textbf{Features/Model} &
  \textbf{Perspective} &
  \textbf{Systems} \\ \addlinespace[0.05cm]\hline \addlinespace[0.05cm]
\multirow{3}{*}{Rule-Based} &
  \multirow{3}{*}{\begin{tabular}[c]{@{}l@{}}Predicate \\ Lexicons\end{tabular}} &
  Author &
  \begin{tabular}[c]{@{}l@{}}\citealp{nairn2006computing} \\ \citealp{lotan2013truthteller}~(TruthTeller)\end{tabular}\\ \addlinespace[0.1cm] 
 &
   &
  Multiple &
  \begin{tabular}[c]{@{}l@{}}\citealp{sauri2012you} \\ (\defacto)\end{tabular} \\ \addlinespace[0.15cm] \hline \addlinespace[0.15cm]
\multirow{8}{*}{\begin{tabular}[c]{@{}l@{}}Feature- \\ Based \\ Supervised \\ Machine \\ Learning\end{tabular}} &
  \multirow{5}{*}{\begin{tabular}[c]{@{}l@{}}Lexico-\\ Syntactic\end{tabular}} &
  Author &
  \begin{tabular}[c]{@{}l@{}}\citealp{diab2009committed}, \citealp{lee2015event} \\ \citealp{prabhakaran2015new}\end{tabular} \\ \addlinespace[0.1cm]
 &
   &
  Reader &
  \begin{tabular}[c]{@{}l@{}}\citealp{de-marneffe-etal-2012-happen}\\ \citealp{soni2014modeling}\end{tabular} \\ \addlinespace[0.1cm]
 &
   &
  Multiple &
  \citealp{qian-2015-ml} \\ \addlinespace[0.1cm] \cline{2-4} \addlinespace[0.1cm] 
 &
  \multirow{2}{*}{\begin{tabular}[c]{@{}l@{}}Output of \\ Rule System\end{tabular}} &
  Author &
  \citealp{stanovsky2017integrating} \\
          &                & Multiple    & \citealp{sauri2012you}              \\ \addlinespace[0.3cm] \hline \addlinespace[0.15cm]
\multirow{3}{*}{\begin{tabular}[c]{@{}l@{}}Neural\\ Networks \\ (NN)\end{tabular}} &
  LSTM &
  Author &
  \citealp{rudinger2018neural} \\
 &
  GAN &
  Multiple &
  \citealp{qian-2018-gan} \\
 &
  Graph NN &
  Author &
  \citealp{pouran-ben-veyseh-etal-2019-graph} \\ \addlinespace[0.15cm] \hline \addlinespace[0.15cm]
\multirow{2}{*}{\begin{tabular}[c]{@{}l@{}}Neural\\ Pretrained\end{tabular}} &
  \multirow{2}{*}{BERT} &
  Author &
  \citealp{jiang2021thinks} \\
 &
   &
  Multiple &
  This work \\ \addlinespace[0.05cm] \hline \addlinespace[0.05cm]
\end{tabular}
\caption{Epistemic stance prediction models.}
\vspace{-0.27cm}
\label{tab:factuality_models}
\end{table}

\paragraph{General Stance Detection in NLP}
Recently, there has been a growing interest in analyzing stance, including a broad spectrum of stance-takers (speaker/writer), the objects of stances, and their relationship. While our work also examines the stance relationship between a source (stance-taker) and an event (object), we differ from the existing literature in several ways. For instance, unlike our work where a stance-taker is the author or a mentioned source in the text, ~\citet{mohtarami-etal-2018-automatic}, \citet{pomerleau2017fake} and \citet{zubiaga-etal-2016-stance} consider the entire document/message to be a stance-taker. Similarly, the object of the stance could be a target entity (such as a person, organization, movement, controversial topic, etc.) that may or may not be explicitly mentioned in the input document~\cite{mohammad-etal-2016-semeval}. On the contrary, in this work, event propositions (object) are always embedded within the text.

Finally, we can also analyze the kind of stance relationship exhibited by the stance-taker towards an object from two linguistic perspectives: affect and epistemic. Affect involves the expression of a broad range of personal attitudes, including emotions, feelings, moods, and general dispositions~\cite{ochs-1989-language}, and has been explored in~\citet{mohammad-etal-2016-semeval}. On the other hand, epistemic---this work's focus---refers to the speaker's expressed attitudes towards knowledge of events and her degree of commitment to the validity of the communicated information~\cite{chafe-1986-evidentiality, biber-1989-styles, palmer-2001-mood}. The analysis explored in ~\citet{mohtarami-etal-2018-automatic}, \citet{pomerleau2017fake} and \citet{zubiaga-etal-2016-stance} seems to be epistemic as they implicitly incorporate the knowledge or claims expressed in the evidence document and hence their stances towards them, although such distinctions are not made explicitly in their work. While the stance literature discussed in this section has not been connected to epistemic stance literature in NLP, we think interesting future work can be done to establish this relationship. 
\section{An Epistemic Stance Framework for Analyzing Political Rhetoric}
\label{sec:epistemic_stance_framework}
This section formally introduces the task of epistemic stance detection and describes the details of the \factbank dataset. We then explain how the epistemic stance framework relates to several rhetorical strategies often used in political discourse.



\subsection{Epistemic Stances}  \label{sec:taskdef}
We define an epistemic stance tuple as a triple of \emph{(source, event, label)} within a sentence, where the label is the value of the source's epistemic stance (or a non-epistemic relation) toward the event.
The triples can be viewed as a fully connected graph among all sources and events in the sentence
(Figure~\ref{fig:motivating_example}).
%
We use the structure and theory of \factbank \cite{sauri2012you} to identify sources, events and the stance labels.

\paragraph{Sources and Events}
A \emph{source} is an entity---either the text's author, or an entity mentioned in the sentence---which can hold beliefs.
\factbank contains annotations for sources that are subjects of source-introducing predicates (SIPs),
a manually curated lexicon of verbs about report and belief such as \emph{claim, doubt, feel, know, say, think}.
Annotations of these embedded sources allow us to analyze the author's depiction of the embedded source's beliefs towards an event. The special \emph{Author} source is additionally included to analyze the author's own beliefs.
\factbank's definition of \emph{events} includes a broad array of textually described eventualities, processes, states, situations, propositions, facts, and possibilities. \factbank identifies its event tokens as those marked in the highly precise, manually annotated \timebank and AQUAINT TimeML\footnote{\scriptsize{\url{https://web.archive.org/web/20070721130754/http://www.timeml.org/site/publications/specs.html}}} corpora.

\begin{figure}
\includegraphics[width=3in, height=1.3in, trim=0in 7.7in 7in 0,clip]{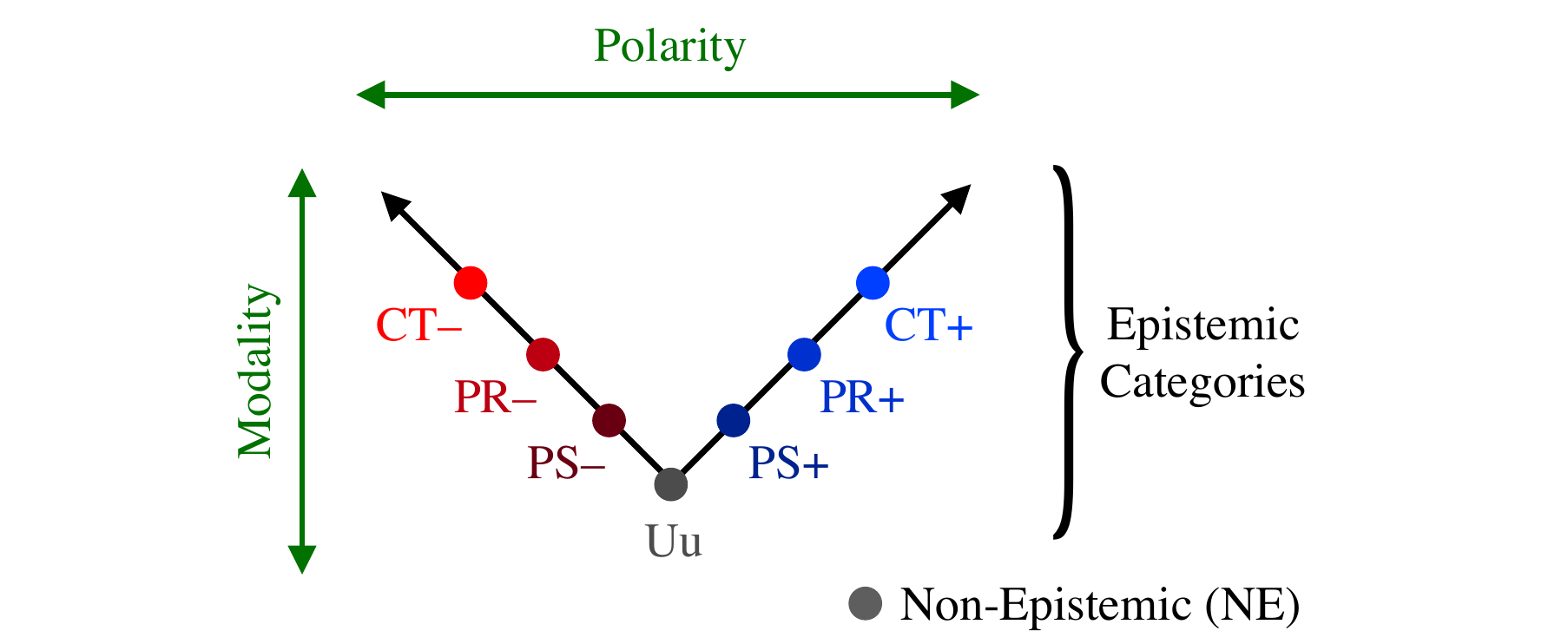}
\caption{Stance labels used in this work, ordered along two linguistic dimensions, as well as a separate non-epistemic category. \label{fig:stance_labels}}
\vspace{-0.27cm}
\end{figure}

\paragraph{Epistemic Stance Label}
\factbank characterizes epistemic stances along two axes, polarity and modality. 
The polarity is binary, with the values positive and negative---the event did (not) happen or the proposition is (not) true.
The modality constitutes a continuum ranging from uncertain to absolutely certain, discretely categorized as \textit{possible (PS)}, \textit{probable (PR)} and \textit{certain (CT)}. An additional \textit{underspecified or uncommitted stance (Uu)} is added along both axes to account for cases such as attribution to another source (non-commitment of the source) or when the stance of the source is unknown. The epistemic stance is then characterized as a pair \textit{(modality, polarity)} containing a modality and a polarity value (e.g., \textit{CT+}) (Figure~\ref{fig:stance_labels}).

FactBank gives epistemic stance labels between certain pairs of sources and events only, based on structural syntactic relationships. However, for raw text we may not have reliable access to syntactic structures, and sources and events must be automatically identified, which may not be completely accurate. We use a simple solution by always assuming edges among the cross-product of all sources and events within a sentence, and to predict a separate \nonepistemic \textit{(NE)} category for the large majority of pairs. This accounts for any spurious event-source pairs, structurally invalid configurations such as an embedded source's stance towards an event outside their factive context (Figure~\ref{fig:motivating_example}: (s4,e2)), or a source that cannot be described as a belief holder (and thus, all its stances are \textit{NE}).

Given that a variety of datasets have been collected for tasks related to epistemic stance (\S\ref{sec:related_work}), \citet{stanovsky2017integrating} argues to combine them for modeling.
However, some datasets address different epistemic questions (e.g.,\  the reader's perspective), and
they follow very different annotation guidelines and annotation strategies, risking ambiguity in labels' meaning.
In preliminary work we attempted to crowdsource new annotations but found the resulting labels to be very different than FactBank, 
which was created by a small group of expert, highly trained annotators.
Thus we decided to exclusively use FactBank for modeling.
\subsection{Connections between Epistemic Stances and Rhetorical Strategies}
\label{sec:epispoli} 
Some epistemic stances in \factbank's framework can be mapped to a common political rhetorical strategy. For instance, a source utilizing \textit{certainly positive/negative} (\textit{CT+/CT-}) stances more frequently can be associated with displaying higher commitment levels. The \textit{CT+/CT-} stances can also help analyze \emph{political disagreements} by identifying two sources with opposite stances towards an event, i.e., a source asserting an event (\textit{CT+}) and a source refuting the same event (\textit{CT-}). A source may exhibit a \textit{probable/possible} (\textit{PR/PS}) stance to indicate that the event could have happened, abstaining from expressing strong commitments towards this event, which can be useful to analyze \emph{hedging}. 
Finally, \textit{underspecified/uncommitted} (\textit{Uu}) stances can help identify the embedded sources whose beliefs are mentioned by the author while remaining uncommitted, a strategy related to \emph{footing-shift} in political discourse. Use of \textit{Uu} stances is also helpful to identify \emph{belief holders}---entities described as having epistemic stances (\S\ref{sec:case_study})---since sometimes the author remains uncommitted while reporting the embedded source's stance.

\vspace{-0.2cm}
\section{Model}
\vspace{-0.2cm}
\label{sec:model}

\begin{table*}[h]
\centering \tiny
\begin{tabular}{lrrrrrrrr}
\hline
\textbf{Model} &
  \multicolumn{1}{c}{\textbf{CT+}} &
  \multicolumn{1}{c}{\textbf{CT-}} &
  \multicolumn{1}{c}{\textbf{PR+}} &
  \multicolumn{1}{c}{\textbf{PS+}} &
  \multicolumn{1}{c}{\textbf{Uu}} &
  \multicolumn{1}{c}{\textbf{NE}} &
  \multicolumn{1}{c}{\textbf{\begin{tabular}[c]{@{}c@{}}Macro Avg\\ (Non-NE)\end{tabular}}} &
  \multicolumn{1}{c}{\textbf{\begin{tabular}[c]{@{}c@{}}Macro Avg\\ (All)\end{tabular}}} \\ \hline
DeFacto~\cite{sauri2012you}            & 85.0 & 75.0 & 46.0 & 59.0  & 75.0 & -    & 70.0 & -    \\
SVM~\cite{sauri2012you, prabhakaran2010automatic}                & 90.0 & 61.0 & 29.0 & 39.0  & 66.0 & -    & 59.0 & -    \\
BiLSTM~\cite{qian-2018-gan}             & 85.2 & 74.0 & 58.2 & 61.3  & 73.3 & -    & 70.4 & -    \\
AC-GAN~\cite{qian-2018-gan}              & 85.5 & 74.1 & 63.1 & 65.4  & 75.1 & -    & 72.6 & -    \\
BERT~\cite{jiang2021thinks}     & 89.7 & 69.8 & 45.0 & 46.7 & 82.8 & 97.9 & 66.8 & 72.0 \\
RoBERTa (this work) & 90.7 & 78.4 & 51.4 & 62.7 & 84.8 & 97.8 & 73.6 & 77.6 \\ \hline
\end{tabular}
\caption{F1 scores for our RoBERTa based epistemic stance classifier and all baseline models.}
\vspace{-0.5cm}
\label{tab:classifier_results}
\end{table*}

We present a simple and reproducible RoBERTa-based neural model for epistemic stance classification using a standard fine-tuning approach.\footnote{We intentionally keep the modeling simple to make it more accessible to political scientists and users with less computational experience.
We further simplify by augmenting \bert with a single task-specific layer, as opposed to a new task-specific model architecture proposed in~\citet{pouran-ben-veyseh-etal-2019-graph, qian-2018-gan, rudinger2018neural}.} BERT fine-tuning is effective for many NLP tasks \cite{devlin-etal-2019-bert}, and recent work on pre-trained language models such as BERT~\cite{shi-etal-2016-string, belinkov2018internal, tenney-etal-2019-bert, tenney2018what, rogers-etal-2020-primer} shows such models encode syntactic and semantic dependencies within a sentence, which is highly related to the epistemic stance task. 

Recently, \citet{jiang2021thinks} use a fine-tuned \bert model for author-only epistemic stance prediction, obtaining strong performance on several datasets. We extend their approach, developing a \bert model (using the RoBERTa~\cite{liu2019roberta} pre-training variant) for the structurally more complex multi-source task, and give the first full comparison to the foundational multi-source system, DeFacto~\cite{sauri2012you}.
We leave the exploration of other advanced transformer-based models~\cite{gpt3, 2020t5} for further performance gains as future work.




To develop a model suitable for multi-source predictions, we follow \citet{tenney2018what} and \citet{rudinger-etal-2018-neural-davidsonian}'s architecture for semantic (proto-role) labeling, which they formulate as predicting labels for pairs of input embeddings.
To predict the epistemic stance for an event-source pair $(e, s)$ in a sentence,
we first compute contextual embeddings for the sentence's tokens, $[h^{L}_{1}, h^{L}_{1},...., h^{L}_{n}]$, 
from a \bert encoder's last ($L^{th}$) layer.
We concatenate the source ($h^{L}_{s}$) and event ($h^{L}_{e}$) token embeddings (each averaged over \bert's sub-token embeddings), and use a single linear layer to parameterize a final softmax prediction $\hat{f} \in [0,1]^{C}$ over the $C=6$ epistemic stance classes,\footnote{CT+, CT-, PR+, PS+, Uu, NE; ~\citet{sauri2012you} additionally define probably/possibly negative (PR-/PS-) stances. However, these stances are rare in the corpus, making modeling and evaluation problematic.
Following \newcite{qian-2015-ml, qian-2018-gan}, we omit them in this study.}
which is trained with cross entropy loss over all tuples in the training set.
We apply inverse frequency class weighting to encourage accurate modeling for comparatively rare classes like the \emph{CT-, PR+ and PS+} class.
Finally, to cleanly analyze the author source in the same manner as other mentioned sources,
we augment the sentence with the prefix ``Author: '' (following a dialogue-like formatting),\footnote{With and without the trailing colon gave same results.} and use its index and embedding for inferences about the author source.

Table~\ref{tab:classifier_results} shows the performance of our RoBERTa based epistemic stance classifier. We compare our model against several baselines, including rule-based methods (DeFacto;~\citet{sauri2012you}), machine learning classifiers (SVM \citet{sauri2012you, prabhakaran2010automatic}), and neural network based methods (BiLSTM and AC-GAN by \citet{qian-2018-gan}) as described in \S\ref{sec:existing_models}.\footnote{Since the DeFacto implementation is not available, we compare our model's predictions on the FactBank test set against evaluation statistics derived from the test set confusion matrix reported by \citeauthor{sauri2012you}. We use implementation provided at \url{https://github.com/qz011/ef_ac_gan} for SVM, BiLSTM and  AC-GAN baselines.} We also extend the author-only BERT model by \citet{jiang2021thinks} to support multi-source predictions in line with our modeling approach. The RoBERTa model performs the best obtaining a macro-averaged F1 score of $77.6\pm$0.011 on all six epistemic labels and an F1 score of $73.6\pm$0.031 on the original five epistemic labels (excluding the \nonepistemic label). Although the RoBERTa model has a much simpler architecture, it performs the same or better than AC-GAN. All pairwise significance tests resulted in $p$-values $<$ $0.01$. Details of implementations and statistical testing is provided in Appendix \S\ref{sec:appendix_implementation_details} and \S\ref{sec:appendix_significance_testing}.

The above epistemic stance classifier, like most previous modeling approaches~\cite{qian-2015-ml, sauri2012you}, 
requires pre-identified sources and events, which do not exist in real-world text.
We use \citet{qian-2018-gan}'s two-step approach
to first identify sources and events in the input text and then determine stances for every recognized (source, event) pair. 
Source and event identification is through two RoBERTa-based token classifiers, using a linear logistic layer for binary classification of whether
a token is a source (or event), fine-tuned on the same training corpus.  

Our source and event identification models achieve a macro-averaged F1 score of $81.8\pm0.019$ and $85.78\pm 0.007$, respectively, slightly improving upon the only existing prior work of ~\citet{qian-2018-gan} by $1.85\%$ and $1.29\%$ respectively, with pairwise significance tests resulting in $p$-values $<$ $0.01$. We also experimented with a joint model to identify sources and events; however, individual classifiers gave us better performance (Appendix \S\ref{sec:appendix_source_event_model}).  
\section{Case Study: Belief Holder Identification}
\label{sec:case_study}
Political discourse involves agreement and contention between the author and other belief-holding sources they cite.
As a first step, we extract major belief holders mentioned in a text to allow
analysis of ideological trends in U.S.\ political discourse.
\subsection{Corpus Description}  \label{sec:caib}
We conduct our case study on the new Mass-Market Manifestos (MMM) corpus, a curated collection
of political nonfiction
authored by U.S.\ politicians, media activists, and opinion elites in English, published from 1993-2020.
It subsumes and more than triples the size of Contemporary American Ideological Books \cite{sim2013measuring}.
The corpus contains 370 books (31.9 million tokens) spanning various U.S. political ideologies.
Human coders identified 133 books as liberal or left-wing, 226 as conservative or right-wing, and 11 as explicitly centrist or independent.
Since ideological opponents often draw from a shared set of concepts---sometimes stating perceived facts and sometimes dismissing others' claims---this presents us with a perfect challenge for detection of epistemic stance.
\subsection{Belief Holder Identification}
\label{sec:bh}
\noindent
A \emph{belief holder} is defined as a non-author source that holds at least one epistemic stance toward some event.
We identify belief holders by using our best-performing model (fine-tuned RoBERTa, predictions averaged over 5 random restarts) to infer epistemic stances for all source-event pairs identified in the $370$ books in the MMM corpus. For the problem of identifying sources that are belief holders as per this definition, we obtain $77.3$ precision and $79.4$ recall on FactBank's evaluation corpus.

For aggregate analysis (\S\ref{sec:polibh}), especially for named entity sources, a longer span is more interpretable and less ambiguous. Thus, when a source token is recognized as part of a traditional named entity
(via spaCy v3.0.6; \citet{honnibal-johnson-2015-improved}),
the belief holder is defined as the full NER span; otherwise, simply the source token is used.
\subsection{Comparison to Named Entity Recognition}  \label{sec:ner}
Instead of using epistemic stance-based belief holder identification,
an alternative approach is to exclusively rely on
named entity recognition (NER) from a set of predefined types.
NER has been used in opinion holder identification \cite{kim-hovy-2004-determining} and within belief evaluation in the TAC KBP Belief/Sentiment track~\cite{tackbp-2016}. 
By contrast, our model can instead find \emph{any} entity as a belief holder, as long as it holds epistemic stances, without a type restriction.
To illustrate this, we compare our belief holder identifier to a standard NER implementation from spaCy v3.0.6~\cite{honnibal-johnson-2015-improved},\footnote{CPU optimized version of \texttt{en$\_$core$\_$web$\_$lg}. Stanza's~\cite{qi2020stanza} performance-optimized NER system gave broadly similar results.}
trained on English web corpus of OntoNotes 5.0~\cite{ontonotes}.
We use entities identified as one of OntoNotes' 11 non-numeric named entity types.\footnote{\emph{Event, Facility, GPE, Language, Law, Location, NORP, Organization, Person, Product, Work\_of\_Art}}
Aggregating among all books in the corpus, the set of belief holders identified by our model
has only a 0.198 Jaccard similarity with the set of NER-detected entities (Appendix \S\ref{sec:appendix_belief_holders} Table~\ref{tab:belief holders} provides qualitative examples from one conservative book).\footnote{An entity is defined as a belief holder if it is the source for at least one epistemic tuple; similarly, it is a named entity if at least one occurrence is identified as part of an NER span.}

\begin{figure}[]
\centering 
\includegraphics[width=0.37\textwidth]{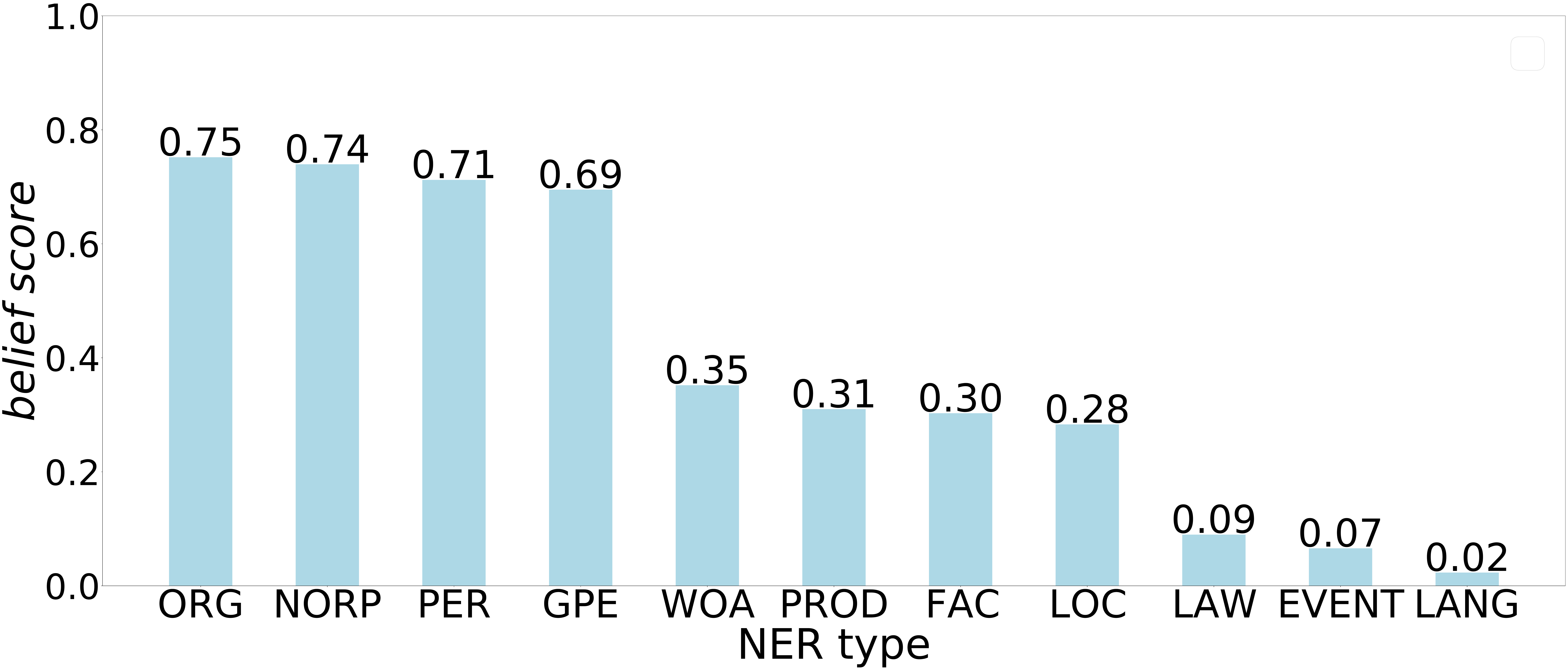}
\caption{Imperfect correlation between belief scores and OntoNotes NER types. {\footnotesize (WOA: Work of Art, PROD: Product, PER: Person, ORG: Organization, LOC: Location, NORP: Nationalities or Religious or Political Groups,
FAC: Facility, LANG: Language, GPE: Geo-Political Entity)}
\vspace{-0.2cm}
\label{fig:ner_dist}} 
\end{figure}

Is it reasonable to define a set of named entity types to identify belief holders?
We calculate each named entity type's \emph{belief score}, which is the average proportion of named entities of that type that are described as holding an epistemic stance.\footnote{For each source instance with same NER type, we find the proportion of epistemic (non-NE) stances among events in its sentence, then average these values across the corpus.}
As shown in Figure~\ref{fig:ner_dist}, while the Organization, NORP, Person and GPE types have significantly higher belief score than others, there is a wide range of variation,
including non-obvious types such as Work of Art (e.g., The Bible), suggesting that a NER type whitelists undercover or overcover possible belief holders. We provide a further linguistic breakdown of identified belief holders in Appendix \S\ref{sec:appendix_linguistic_analysis}.

\begin{table}[t]
\centering
\tiny
\begin{tabular}{lclc}

\multicolumn{2}{c}{{Highly Cited by Left-wing Authors}}  &
\multicolumn{2}{c}{{Highly Cited by Right-wing Authors}}  \\
\cmidrule(lr){1-2} \cmidrule(lr){3-4}

\multicolumn{1}{c}{{Belief Holder}} &
\multicolumn{1}{c}{{View}} &
\multicolumn{1}{c}{{Belief Holder}} &
\multicolumn{1}{c}{{View}} \\
\cmidrule(lr){1-1}\cmidrule(lr){2-2}\cmidrule(lr){3-3}\cmidrule(lr){4-4}
Tom Delay          & Opposed   & Paul Johnson          & Respected \\
Martin Gilens      & Respected   & Marvin Olasky         & Respected          \\
Michelle Alexander & Respected & Saul Alinsky          & Opposed   \\
Grover Norquist    & Opposed   & Robert Rector         & Respected          \\
Jane Mayer         & Respected & Thomas Sowell         & Respected          \\
Albert Camus       & Respected  & The Tax Foundation    & Respected          \\
Consumers          &   Respected       & Soviets               & Opposed   \\
Thomas Edsall      &   Respected        & George Soros          & Opposed   \\
Jacob Hacker       & Respected & Pew Research          & Respected \\
James Baldwin      & Respected & John Edwards          & Opposed   \\
Jeffrey Sachs      & Respected & George Stephanopoulos & Opposed   \\
Michele Bachmann   & Opposed   & John Stossel          & Respected \\
Ben Bernanke       &  Unclear         & Thomas Sowell         & Respected \\
Chris Hedges       & Respected & Nicholas Eberstadt    & Respected          \\
Lobbyists          & Opposed   & James Wilson          & Respected \\
Bill Moyers        & Respected & Iran                  & Opposed   \\
Daniel Bell        & Respected & Hollywood             & Opposed   \\
David Cay Johnston & Respected & George Gilder         & Respected \\
Instructor         &  Generic         & Dennis Prager         & Respected \\
Moderator          &  Generic         & Arthur Brooks         & Respected          \\ 
\end{tabular}
\caption{Top 20 most frequently mentioned belief holders per author ideology (left vs.\ right),
among belief holders mentioned in $\geq$ 8 books in the MMM corpus.}
\vspace{-0.6cm}
\label{table:belief_holders_8}
\end{table}

\subsection{Political Analysis of Belief Holders}
\label{sec:polibh}
The MMM corpus, including both left and right-wing authors,
gives an opportunity to study the belief holder citation practices for each U.S.\ political ideology.
Using our epistemic stance and entity aggregation postprocessing (\S\ref{sec:bh}),
we count the number of books each belief holder is mentioned in.
There are 1269 sources mentioned as a belief holder in $\geq$ 8 books.
For each belief holder, we calculate its left-right citation ratio:
the proportion of left-wing books it is mentioned in, versus the proportion of right-wing books (proportions are calculated using a book pseudocount of 1 to avoid dividing by zero).
Belief holders with a ratio $\sim$ 1.0 include some generic (\emph{team, organization, official}) and
anaphoric (\emph{anyone, many}) examples.



Table~\ref{table:belief_holders_8} shows the top 20 belief holders for both left and right, as ranked by this ratio, yielding a rich set of politicians (Delay, Edwards), journalists (Mayer, Stephanopoulos), activists (Norquist, Alinsky),
and many social scientists and scholars (Gilens, Johnson).
Most of these belief holders were recognized by an expert (political scientist coauthor) as being respected or opposed from the citing ideological perspective. 
Based on prior knowledge of U.S.\ politics it was straightforward to immediately give such judgments for most entries; for a few unclear ones, we checked individual sentences mentioning the belief holder.
A common strategy is to describe an opponent's views or statements---the use of a rhetorical bogeyman.

\begin{table}[]
\centering
\tiny
\begin{tabular}{lllll}
\multicolumn{2}{c}{Left-cited} &  & \multicolumn{2}{c}{Right-cited} \\ \cline{1-2} \cline{4-5} 
Economists    & Studies        &  & Founders      & Democrats       \\
Woman         & Research       &  & Media         & Officials       \\
Polls         & Republicans    &  & Poll          & President       \\
Scientists    & Group          &  & Obama         & Conservatives   \\
Groups        & Friend         &  & Government    & Liberals       
\end{tabular}
\caption{Top 10 most frequently mentioned belief holders per author ideology, among belief holders mentioned in at least 100 books.}
\vspace{-0.2cm}
\label{table:belief_holders_100}
\end{table}

\begin{figure}
\centering
  \fbox{
  \centering
  \begin{minipage}[t]{0.4\textwidth}
    \centering
    \scriptsize
    \begin{itemizesquish}
        \item 
We know that most of the \textbf{[Founders]}$_{s}$ regarded slavery as a wrong that would have to be addressed. \emph{Chuck Norris, Black Belt Patriotism (R)}
\item
Sometimes, whether against gator or human predator, you're on your own, as the frontier-expanding \textbf{[Founders]}$_{s}$ well knew. \emph{Charlie Kirk, The MAGA Doctrine (R)}
\item
This is not to say the \textbf{[founders]}$_{s}$ believed that only religious individuals could possess good character. \emph{William Bennett, America the Strong (R)}
\item
The \textbf{[founders]}$_{s}$, however, had quite another idea, based on their experience in the colonies over the decades before, where actual religious freedom had existed. \emph{Eric Metaxas, If You Can Keep It (R)}
\item
The \textbf{[Founders]}$_{s}$ recognized that there were seeds of anarchy in the idea of individual freedom [..], for if everybody is truly free, without the constraints of birth or rank or an inherited social order [..] then how can we ever hope to form a society that coheres?
\emph{Barack Obama, The Audacity of Hope (L)}

\end{itemizesquish}

\end{minipage}
}

\caption{Examples of \emph{founders} as a belief holder.}
\vspace{-0.5cm}
\label{fig:founders_examples}
\end{figure}

Repeating the analysis for widely cited belief holders
appearing in $\geq$ 100 books, yields more general, and again politically meaningful, entities (Table~\ref{table:belief_holders_100}).
Some well-known patterns are clearly visible, such as
liberals' respect for technocratic authority (\emph{economists, scientists, research}),
and conservative respect for the semi-mythical \emph{founders} alongside derision for the \emph{media}.
Both sides frequently cite the opposition (L: \emph{Republicans}, R: \emph{Democrats}),
though interestingly the right cites both conservatives and liberals (relatively more frequently than the left).
Figure~\ref{fig:founders_examples} shows examples of \emph{founders}, with the most skewed ratio ($0.308 \approx 3.2^{-1}$) among this set of entities.
Overall, our automated belief holder identification yields a politically significant entity list, laying the groundwork for more systematic manual and computational analysis (e.g., network or targeted sentiment analysis).
\vspace{-0.15cm}
\section{Conclusion}
\vspace{-0.15cm}
Semantic modeling has exciting potential to deepen the NLP analysis of political discourse. In this work, we analyze the epistemic stance of various sources toward events, by developing a RoBERTa-based model, and using it for identifying major belief holders mentioned by political authors. We conduct a large-scale analysis of the Mass Market Manifestos corpus of U.S.\ political opinion books, where we characterize trends in cited belief holders across U.S.\ political ideologies. In future, we hope to use this framework to help construct a database of beliefs, belief holders, and their patterns of agreement and disagreement in contentious domains.

\section*{Acknowledgements}
We are thankful for the feedback and comments from the reviewers. We are grateful to Philip Resnik, Michael Colaresi, Arthur Spirling, Katherine Keith, Geraud Nangue Tasse, Kalpesh Krishna, Marzena Karpinska, and the UMass NLP group for valuable discussions during the course of the project. This work was supported by National Science Foundation grants 1814955 and 1845576. Any opinions, findings, conclusions or recommendations expressed in this material are those of the authors and do not necessarily reflect the views of the National Science Foundation.

\bibliography{custom}

\begin{thebibliography}{111}
\expandafter\ifx\csname natexlab\endcsname\relax\def\natexlab#1{#1}\fi

\bibitem[{Abdul-Jabbar and Obstfeld(2016)}]{abdul-2016-writings}
Kareem Abdul-Jabbar and Raymond Obstfeld. 2016.
\newblock \emph{Writings on the wall: Searching for a new equality beyond Black
  and White}.
\newblock Time Inc. Books.

\bibitem[{Anderson(1986)}]{anderson-1986-evidentials}
Lloyd~B. Anderson. 1986.
\newblock Evidentials, paths of change, and mental maps: Typologically regular
  asymmetries.
\newblock In \emph{Wallace L. Chafe and Johanna Nichols (eds.), Evidentiality:
  The Linguistic Coding of Epistemology}, pages 273--312. Ablex.

\bibitem[{Arrese(2009)}]{arrese-2009-effective}
Juana I~Mar\'{i}n Arrese. 2009.
\newblock Effective vs. {E}pistemic stance, and subjectivity/intersubjectivity
  in political discourse. {A} case study.
\newblock \emph{Studies on English modality: {I}n honour of Frank Palmer.
  Linguistic Insights}, 111:23--131.

\bibitem[{Beck(2018)}]{beck2018addicted}
Glenn Beck. 2018.
\newblock \emph{Addicted to outrage: How thinking like a recovering addict can
  heal the country}.
\newblock Threshold Editions.

\bibitem[{Belinkov(2018)}]{belinkov2018internal}
Yonatan Belinkov. 2018.
\newblock \href {http://hdl.handle.net/1721.1/118079} {\emph{On internal
  language representations in Deep Learning: An analysis of Machine Translation
  and Speech Recognition}}.
\newblock Ph.D. thesis, Massachusetts Institute of Technology, Cambridge,
  {USA}.

\bibitem[{Berg-Kirkpatrick et~al.(2012)Berg-Kirkpatrick, Burkett, and
  Klein}]{berg-kirkpatrick-etal-2012-empirical}
Taylor Berg-Kirkpatrick, David Burkett, and Dan Klein. 2012.
\newblock \href {https://aclanthology.org/D12-1091} {An empirical investigation
  of statistical significance in {NLP}}.
\newblock In \emph{Proceedings of the 2012 Joint Conference on Empirical
  Methods in Natural Language Processing and Computational Natural Language
  Learning}, pages 995--1005, Jeju Island, Korea. Association for Computational
  Linguistics.

\bibitem[{Bethard et~al.(2004)Bethard, Yu, Thornton, Hatzivassiloglou, and
  Jurafsky}]{bethard2004automatic}
Steven Bethard, Hong Yu, Ashley Thornton, Vasileios Hatzivassiloglou, and Dan
  Jurafsky. 2004.
\newblock \href {https://web.stanford.edu/~jurafsky/SS404BethardS.pdf}
  {Automatic extraction of opinion propositions and their holders}.
\newblock In \emph{Proceedings of the AAAI Spring Symposium on Exploring
  Attitude and Affect in Text: Theories and Applications}, volume 2224.

\bibitem[{Biber and Finegan(1989)}]{biber-1989-styles}
Douglas Biber and Edward Finegan. 1989.
\newblock \href {https://doi.org/10.1515/text.1.1989.9.1.93} {Styles of stance
  in english: Lexical and grammatical marking of evidentiality and affect}.
\newblock \emph{Text-Interdisciplinary Journal for the study of Discourse},
  9(1):93--124.

\bibitem[{Biran et~al.(2012)Biran, Rosenthal, Andreas, McKeown, and
  Rambow}]{biran2012detecting}
Or~Biran, Sara Rosenthal, Jacob Andreas, Kathleen McKeown, and Owen Rambow.
  2012.
\newblock Detecting influencers in written online conversations.

\bibitem[{Bracewell et~al.(2012)Bracewell, Tomlinson, and
  Wang}]{bracewell2012motif}
David~B Bracewell, Marc Tomlinson, and Hui Wang. 2012.
\newblock A motif approach for identifying pursuits of power in social
  discourse.
\newblock In \emph{2012 IEEE Sixth International Conference on Semantic
  Computing}, pages 1--8. IEEE.

\bibitem[{Brown et~al.(2020)Brown, Mann, Ryder, Subbiah, Kaplan, Dhariwal,
  Neelakantan, Shyam, Sastry, Askell, Agarwal, Herbert-Voss, Krueger, Henighan,
  Child, Ramesh, Ziegler, Wu, Winter, Hesse, Chen, Sigler, Litwin, Gray, Chess,
  Clark, Berner, McCandlish, Radford, Sutskever, and Amodei}]{gpt3}
Tom Brown, Benjamin Mann, Nick Ryder, Melanie Subbiah, Jared~D Kaplan, Prafulla
  Dhariwal, Arvind Neelakantan, Pranav Shyam, Girish Sastry, Amanda Askell,
  Sandhini Agarwal, Ariel Herbert-Voss, Gretchen Krueger, Tom Henighan, Rewon
  Child, Aditya Ramesh, Daniel Ziegler, Jeffrey Wu, Clemens Winter, Chris
  Hesse, Mark Chen, Eric Sigler, Mateusz Litwin, Scott Gray, Benjamin Chess,
  Jack Clark, Christopher Berner, Sam McCandlish, Alec Radford, Ilya Sutskever,
  and Dario Amodei. 2020.
\newblock \href
  {https://proceedings.neurips.cc/paper/2020/file/1457c0d6bfcb4967418bfb8ac142f64a-Paper.pdf}
  {Language models are few-shot learners}.
\newblock In \emph{Advances in Neural Information Processing Systems},
  volume~33, pages 1877--1901. Curran Associates, Inc.

\bibitem[{Buchanan(1999)}]{pat1999republic}
Pat Buchanan. 1999.
\newblock \emph{A Republic, Not an Empire: Reclaiming {A}merica's Destiny}.
\newblock Regnery Publishing.

\bibitem[{Chafe(1986)}]{chafe-1986-evidentiality}
Wallace Chafe. 1986.
\newblock Evidentiality in {E}nglish conversation and academic writing.
\newblock In \emph{W. Chafe, \& J. Nichols (eds.), Evidentiality: The
  Linguistic Coding of Epistemology}, pages 261--272. Ablex.

\bibitem[{Chilton(2004)}]{chilton-2004-analysing}
Paul Chilton. 2004.
\newblock \href {https://doi.org/https://doi.org/10.4324/9780203561218}
  {\emph{Analysing Political Discourse: Theory and Practice}}.
\newblock London: Routledge.

\bibitem[{Choi et~al.(2005)Choi, Cardie, Riloff, and
  Patwardhan}]{choi-2005-identifying}
Yejin Choi, Claire Cardie, Ellen Riloff, and Siddharth Patwardhan. 2005.
\newblock \href {https://doi.org/10.3115/1220575.1220620} {Identifying sources
  of opinions with {C}onditional {R}andom {F}ields and extraction patterns}.
\newblock In \emph{Proceedings of the Conference on Human Language Technology
  and Empirical Methods in Natural Language Processing}, HLT '05, page
  355–362, USA. Association for Computational Linguistics.

\bibitem[{Christensen(2009)}]{christensen2009disagreement}
David Christensen. 2009.
\newblock Disagreement as evidence: {T}he epistemology of controversy.
\newblock \emph{Philosophy Compass}, 4(5):756--767.

\bibitem[{Clayman(1992)}]{clayman1992footing}
Steven~E. Clayman. 1992.
\newblock Footing in the achievement of neutrality: {T}he case of news
  interview discourse.
\newblock In \emph{Paul Drew and John Heritage (eds.), Talk at Work:
  Interaction in Institutional Settings}, page 163–198. Cambridge University
  Press, Cambridge.

\bibitem[{Coulter(2009)}]{coulter2009guilty}
Ann Coulter. 2009.
\newblock Guilty: Liberal ``victims" and their assault on {A}merica.
\newblock Crown Forum.

\bibitem[{Danescu-Niculescu-Mizil et~al.(2012)Danescu-Niculescu-Mizil, Lee,
  Pang, and Kleinberg}]{danescu2012echoes}
Cristian Danescu-Niculescu-Mizil, Lillian Lee, Bo~Pang, and Jon Kleinberg.
  2012.
\newblock Echoes of power: Language effects and power differences in social
  interaction.
\newblock In \emph{Proceedings of the 21st international conference on World
  Wide Web}, pages 699--708.

\bibitem[{de~Marneffe et~al.(2012)de~Marneffe, Manning, and
  Potts}]{de-marneffe-etal-2012-happen}
Marie-Catherine de~Marneffe, Christopher~D. Manning, and Christopher Potts.
  2012.
\newblock \href {https://doi.org/10.1162/COLI_a_00097} {Did it happen? the
  pragmatic complexity of veridicality assessment}.
\newblock \emph{Computational Linguistics}, 38(2):301--333.

\bibitem[{Devlin et~al.(2019)Devlin, Chang, Lee, and
  Toutanova}]{devlin-etal-2019-bert}
Jacob Devlin, Ming-Wei Chang, Kenton Lee, and Kristina Toutanova. 2019.
\newblock \href {https://doi.org/10.18653/v1/N19-1423} {{BERT}: Pre-training of
  deep bidirectional transformers for language understanding}.
\newblock In \emph{Proceedings of the 2019 Conference of the North {A}merican
  Chapter of the Association for Computational Linguistics: Human Language
  Technologies, Volume 1 (Long and Short Papers)}, pages 4171--4186,
  Minneapolis, Minnesota. Association for Computational Linguistics.

\bibitem[{Dhingra et~al.(2019)Dhingra, Faruqui, Parikh, Chang, Das, and
  Cohen}]{dhingra2019handling}
Bhuwan Dhingra, Manaal Faruqui, Ankur Parikh, Ming-Wei Chang, Dipanjan Das, and
  William Cohen. 2019.
\newblock \href {https://doi.org/10.18653/v1/P19-1483} {Handling divergent
  reference texts when evaluating table-to-text generation}.
\newblock In \emph{Proceedings of the 57th Annual Meeting of the Association
  for Computational Linguistics}, pages 4884--4895, Florence, Italy.
  Association for Computational Linguistics.

\bibitem[{Diab et~al.(2009)Diab, Levin, Mitamura, Rambow, Prabhakaran, and
  Guo}]{diab2009committed}
Mona Diab, Lori Levin, Teruko Mitamura, Owen Rambow, Vinodkumar Prabhakaran,
  and Weiwei Guo. 2009.
\newblock \href {https://aclanthology.org/W09-3012} {Committed belief
  annotation and tagging}.
\newblock In \emph{Proceedings of the Third Linguistic Annotation Workshop
  ({LAW} {III})}, pages 68--73, Suntec, Singapore. Association for
  Computational Linguistics.

\bibitem[{Dodge et~al.(2020)Dodge, Ilharco, Schwartz, Farhadi, Hajishirzi, and
  Smith}]{dodge-jesse-etal-2020-fine}
Jesse Dodge, Gabriel Ilharco, Roy Schwartz, Ali Farhadi, Hannaneh Hajishirzi,
  and Noah~A. Smith. 2020.
\newblock \href {https://arxiv.org/abs/2002.06305} {Fine-tuning pretrained
  language models: Weight initializations, data orders, and early stopping}.
\newblock \emph{Computing Research Repository}, arXiv:2002.06305.

\bibitem[{Dreher(2018)}]{dreher2018benedict}
Rod Dreher. 2018.
\newblock \emph{The Benedict Option: A strategy for {C}hristians in a
  post-{C}hristian nation}.
\newblock Penguin.

\bibitem[{Forbes and Ames(2012)}]{forbes-2012-freedom}
Steve Forbes and Elizabeth Ames. 2012.
\newblock \emph{Freedom Manifesto: Why free markets are moral and big
  government isn't}.
\newblock Currency.

\bibitem[{Frances(2014)}]{frances2014disagreement}
Bryan Frances. 2014.
\newblock \emph{Disagreement}.
\newblock John Wiley \& Sons.

\bibitem[{Fraser(2010)}]{fraser2010hedging}
Bruce Fraser. 2010.
\newblock Hedging in political discourse.
\newblock \emph{Perspectives in Politics and Discourse}, pages 201--214.

\bibitem[{Gardner et~al.(2018)Gardner, Grus, Neumann, Tafjord, Dasigi, Liu,
  Peters, Schmitz, and Zettlemoyer}]{gardner-etal-2018-allennlp}
Matt Gardner, Joel Grus, Mark Neumann, Oyvind Tafjord, Pradeep Dasigi,
  Nelson~F. Liu, Matthew Peters, Michael Schmitz, and Luke Zettlemoyer. 2018.
\newblock \href {https://doi.org/10.18653/v1/W18-2501} {{A}llen{NLP}: A deep
  semantic natural language processing platform}.
\newblock In \emph{Proceedings of Workshop for {NLP} Open Source Software
  ({NLP}-{OSS})}, pages 1--6, Melbourne, Australia. Association for
  Computational Linguistics.

\bibitem[{Goffman(1981)}]{goffman1981forms}
Erving Goffman. 1981.
\newblock \emph{Forms of talk}.
\newblock University of Pennsylvania Press.

\bibitem[{Heider(1946)}]{heider1946attitudes}
Fritz Heider. 1946.
\newblock Attitudes and cognitive organization.
\newblock \emph{The Journal of psychology}, 21(1):107--112.

\bibitem[{Hintikka(1962)}]{hintikka1962knowledge}
Kaarlo Jaakko~Juhani Hintikka. 1962.
\newblock Knowledge and {B}elief: {A}n introduction to the logic of the two
  notions.
\newblock \emph{Cornell University Press}.

\bibitem[{Hoek(1990)}]{hoek1990systems}
Wiebe van~der Hoek. 1990.
\newblock Systems for knowledge and beliefs.
\newblock In \emph{European Workshop on Logics in Artificial Intelligence},
  pages 267--281. Springer.

\bibitem[{Holliday(2018)}]{holliday2018epistemic}
Wesley~H Holliday. 2018.
\newblock Epistemic {L}ogic and {E}pistemology.
\newblock In \emph{Introduction to Formal Philosophy}, pages 351--369.
  Springer.

\bibitem[{Honnibal and Johnson(2015)}]{honnibal-johnson-2015-improved}
Matthew Honnibal and Mark Johnson. 2015.
\newblock \href {https://doi.org/10.18653/v1/D15-1162} {An improved
  non-monotonic transition system for dependency parsing}.
\newblock In \emph{Proceedings of the 2015 Conference on Empirical Methods in
  Natural Language Processing}, pages 1373--1378, Lisbon, Portugal. Association
  for Computational Linguistics.

\bibitem[{Horn(1972)}]{horn1972semantic}
Laurence~Robert Horn. 1972.
\newblock \emph{On the semantic properties of logical operators in English}.
\newblock University of California, Los Angeles.

\bibitem[{Hovy et~al.(2006)Hovy, Marcus, Palmer, Ramshaw, and
  Weischedel}]{ontonotes}
Eduard Hovy, Mitchell Marcus, Martha Palmer, Lance Ramshaw, and Ralph
  Weischedel. 2006.
\newblock \href {https://aclanthology.org/N06-2015} {{O}nto{N}otes: The 90{\%}
  solution}.
\newblock In \emph{Proceedings of the Human Language Technology Conference of
  the {NAACL}, Companion Volume: Short Papers}, pages 57--60, New York City,
  USA. Association for Computational Linguistics.

\bibitem[{Hyland(1996)}]{hyland1996writing}
Ken Hyland. 1996.
\newblock Writing without conviction? {H}edging in science research articles.
\newblock \emph{Applied Linguistics}, 17(4):433--454.

\bibitem[{Jalilifar and Alavi(2011)}]{jalilifar2011power}
Ali~Reza Jalilifar and Maryam Alavi. 2011.
\newblock Power and {P}olitics of language use: {A} survey of hedging devices
  in political interviews.
\newblock \emph{Journal of Teaching Language Skills}, 30(3):43--66.

\bibitem[{Jang(2009)}]{jang2009diverse}
Seung-Jin Jang. 2009.
\newblock Are diverse political networks always bad for participatory
  democracy? {I}ndifference, alienation, and political disagreements.
\newblock \emph{American Politics Research}, 37(5):879--898.

\bibitem[{Jiang and de~Marneffe(2021)}]{jiang2021thinks}
Nanjiang Jiang and Marie{-}Catherine de~Marneffe. 2021.
\newblock \href {http://arxiv.org/abs/2107.00807} {He thinks he knows better
  than the doctors: {BERT} for event factuality fails on pragmatics}.
\newblock \emph{CoRR}, cs.CL/2107.00807v1.

\bibitem[{Kiefer(1987)}]{kiefer1987defining}
Ferenc Kiefer. 1987.
\newblock On defining modality.
\newblock \emph{Walter de Gruyter}.

\bibitem[{Kim and Hovy(2004)}]{kim-hovy-2004-determining}
Soo-Min Kim and Eduard Hovy. 2004.
\newblock \href {https://aclanthology.org/C04-1200} {Determining the sentiment
  of opinions}.
\newblock In \emph{{COLING} 2004: Proceedings of the 20th International
  Conference on Computational Linguistics}, pages 1367--1373, Geneva,
  Switzerland. COLING.

\bibitem[{Kingma and Ba(2015)}]{DBLP:journals/corr/KingmaB14}
Diederik~P. Kingma and Jimmy Ba. 2015.
\newblock \href {http://arxiv.org/abs/1412.6980} {Adam: {A} method for
  stochastic optimization}.
\newblock In \emph{3rd International Conference on Learning Representations,
  {ICLR} 2015, San Diego, CA, USA, May 7-9, 2015, Conference Track
  Proceedings}.

\bibitem[{Klofstad et~al.(2013)Klofstad, Sokhey, and
  McClurg}]{klofstad2013disagreeing}
Casey~A Klofstad, Anand~Edward Sokhey, and Scott~D McClurg. 2013.
\newblock Disagreeing about disagreement: How conflict in social networks
  affects political behavior.
\newblock \emph{American Journal of Political Science}, 57(1):120--134.

\bibitem[{Kryscinski et~al.(2019)Kryscinski, Keskar, McCann, Xiong, and
  Socher}]{kryscinski-etal-2019-neural}
Wojciech Kryscinski, Nitish~Shirish Keskar, Bryan McCann, Caiming Xiong, and
  Richard Socher. 2019.
\newblock \href {https://doi.org/10.18653/v1/D19-1051} {Neural text
  summarization: A critical evaluation}.
\newblock In \emph{Proceedings of the 2019 Conference on Empirical Methods in
  Natural Language Processing and the 9th International Joint Conference on
  Natural Language Processing (EMNLP-IJCNLP)}, pages 540--551, Hong Kong,
  China. Association for Computational Linguistics.

\bibitem[{Lakoff(1975)}]{lakoff1975hedges}
George Lakoff. 1975.
\newblock Hedges: A study in meaning criteria and the logic of fuzzy concepts.
\newblock In \emph{Contemporary research in philosophical logic and linguistic
  semantics}, pages 221--271. Springer.

\bibitem[{Langacker(2009)}]{langacker-2009-investigations}
Ronald~W. Langacker. 2009.
\newblock \href {https://doi.org/doi:10.1515/9783110214369}
  {\emph{Investigations in {C}ognitive {G}rammar}}.
\newblock De Gruyter Mouton.

\bibitem[{Lee et~al.(2015)Lee, Artzi, Choi, and Zettlemoyer}]{lee2015event}
Kenton Lee, Yoav Artzi, Yejin Choi, and Luke Zettlemoyer. 2015.
\newblock \href {https://doi.org/10.18653/v1/D15-1189} {Event detection and
  factuality assessment with non-expert supervision}.
\newblock In \emph{Proceedings of the 2015 Conference on Empirical Methods in
  Natural Language Processing}, pages 1643--1648, Lisbon, Portugal. Association
  for Computational Linguistics.

\bibitem[{Liu(2012)}]{liu2012sentiment}
Bing Liu. 2012.
\newblock \href
  {https://doi.org/https://doi.org/10.2200/S00416ED1V01Y201204HLT016}
  {Sentiment analysis and opinion mining}.
\newblock \emph{Synthesis lectures on {H}uman {L}anguage {T}echnologies},
  5(1):1--167.

\bibitem[{Liu et~al.(2019)Liu, Ott, Goyal, Du, Joshi, Chen, Levy, Lewis,
  Zettlemoyer, and Stoyanov}]{liu2019roberta}
Yinhan Liu, Myle Ott, Naman Goyal, Jingfei Du, Mandar Joshi, Danqi Chen, Omer
  Levy, Mike Lewis, Luke Zettlemoyer, and Veselin Stoyanov. 2019.
\newblock Roberta: A robustly optimized {BERT} pretraining approach.
\newblock \emph{arXiv preprint arXiv:1907.11692}.

\bibitem[{Lotan et~al.(2013)Lotan, Stern, and Dagan}]{lotan2013truthteller}
Amnon Lotan, Asher Stern, and Ido Dagan. 2013.
\newblock \href {https://aclanthology.org/N13-1091} {{T}ruth{T}eller:
  Annotating predicate truth}.
\newblock In \emph{Proceedings of the 2013 Conference of the North {A}merican
  Chapter of the Association for Computational Linguistics: Human Language
  Technologies}, pages 752--757, Atlanta, Georgia. Association for
  Computational Linguistics.

\bibitem[{Lyons(1977)}]{lyons1977semantics}
John Lyons. 1977.
\newblock Semantics.
\newblock \emph{Cambridge University Press}.

\bibitem[{MacKinnon(2009)}]{mackinnon2009bootstrap}
James~G MacKinnon. 2009.
\newblock Bootstrap hypothesis testing.
\newblock \emph{Handbook of Computational Econometrics}, 183:213.

\bibitem[{Martin and White(2005)}]{martin2005appraisal}
James~R Martin and Peter R~R White. 2005.
\newblock The language of evaluation. {A}ppraisal in {E}nglish.
\newblock \emph{Palgrave Macmillan London}.

\bibitem[{Maynez et~al.(2020)Maynez, Narayan, Bohnet, and
  McDonald}]{maynez-etal-2020-faithfulness}
Joshua Maynez, Shashi Narayan, Bernd Bohnet, and Ryan McDonald. 2020.
\newblock \href {https://doi.org/10.18653/v1/2020.acl-main.173} {On
  faithfulness and factuality in abstractive summarization}.
\newblock In \emph{Proceedings of the 58th Annual Meeting of the Association
  for Computational Linguistics}, pages 1906--1919, Online. Association for
  Computational Linguistics.

\bibitem[{Mihaylova et~al.(2018)Mihaylova, Nakov, M{\`a}rquez,
  Barr{\'o}n-Cede{\~n}o, Mohtarami, Karadzhov, and Glass}]{mihaylova2018fact}
Tsvetomila Mihaylova, Preslav Nakov, Llu{\'\i}s M{\`a}rquez, Alberto
  Barr{\'o}n-Cede{\~n}o, Mitra Mohtarami, Georgi Karadzhov, and James Glass.
  2018.
\newblock \href {https://ojs.aaai.org/index.php/AAAI/article/view/11983} {Fact
  checking in community forums}.
\newblock \emph{Proceedings of the AAAI Conference on Artificial Intelligence},
  32(1).

\bibitem[{Minard et~al.(2016)Minard, Speranza, Urizar, Altuna, van Erp, Schoen,
  and van Son}]{minard2016meantime}
Anne-Lyse Minard, Manuela Speranza, Ruben Urizar, Bego{\~n}a Altuna, Marieke
  van Erp, Anneleen Schoen, and Chantal van Son. 2016.
\newblock \href {https://aclanthology.org/L16-1699} {{MEANTIME}, the
  {N}ews{R}eader multilingual event and time corpus}.
\newblock In \emph{Proceedings of the Tenth International Conference on
  Language Resources and Evaluation ({LREC}'16)}, pages 4417--4422,
  Portoro{\v{z}}, Slovenia. European Language Resources Association (ELRA).

\bibitem[{Mohammad et~al.(2016)Mohammad, Kiritchenko, Sobhani, Zhu, and
  Cherry}]{mohammad-etal-2016-semeval}
Saif Mohammad, Svetlana Kiritchenko, Parinaz Sobhani, Xiaodan Zhu, and Colin
  Cherry. 2016.
\newblock \href {https://doi.org/10.18653/v1/S16-1003} {{S}em{E}val-2016 task
  6: Detecting stance in tweets}.
\newblock In \emph{Proceedings of the 10th International Workshop on Semantic
  Evaluation ({S}em{E}val-2016)}, pages 31--41, San Diego, California.
  Association for Computational Linguistics.

\bibitem[{Mohtarami et~al.(2018)Mohtarami, Baly, Glass, Nakov, M{\`a}rquez, and
  Moschitti}]{mohtarami-etal-2018-automatic}
Mitra Mohtarami, Ramy Baly, James Glass, Preslav Nakov, Llu{\'\i}s M{\`a}rquez,
  and Alessandro Moschitti. 2018.
\newblock \href {https://doi.org/10.18653/v1/N18-1070} {Automatic stance
  detection using end-to-end memory networks}.
\newblock In \emph{Proceedings of the 2018 Conference of the North {A}merican
  Chapter of the Association for Computational Linguistics: Human Language
  Technologies, Volume 1 (Long Papers)}, pages 767--776, New Orleans,
  Louisiana. Association for Computational Linguistics.

\bibitem[{Mosbach et~al.(2021)Mosbach, Andriushchenko, and
  Klakow}]{mosbach2021on}
Marius Mosbach, Maksym Andriushchenko, and Dietrich Klakow. 2021.
\newblock \href {https://openreview.net/forum?id=nzpLWnVAyah} {On the stability
  of fine-tuning {BERT}: Misconceptions, explanations, and strong baselines}.
\newblock In \emph{International Conference on Learning Representations}.

\bibitem[{Mushin(2001)}]{mushin2001evidentiality}
Ilana Mushin. 2001.
\newblock Evidentiality and {E}pistemological stance.
\newblock \emph{Narrative Retelling}.

\bibitem[{Nairn et~al.(2006)Nairn, Condoravdi, and
  Karttunen}]{nairn2006computing}
Rowan Nairn, Cleo Condoravdi, and Lauri Karttunen. 2006.
\newblock \href {https://aclanthology.org/W06-3907} {Computing relative
  polarity for textual inference}.
\newblock In \emph{Proceedings of the Fifth International Workshop on Inference
  in Computational Semantics ({IC}o{S}-5)}.

\bibitem[{Ochs and Schieffelin(1989)}]{ochs-1989-language}
Elinor Ochs and Bambi Schieffelin. 1989.
\newblock \href {https://doi.org/doi:10.1515/text.1.1989.9.1.7} {Language has a
  heart}.
\newblock \emph{Text - Interdisciplinary Journal for the Study of Discourse},
  9(1):7--26.

\bibitem[{Palmer(2001)}]{palmer-2001-mood}
Frank~Robert Palmer. 2001.
\newblock \emph{Mood and {M}odality}.
\newblock Cambridge University Press.

\bibitem[{Pang and Lee(2008)}]{pang2009opinion}
Bo~Pang and Lillian Lee. 2008.
\newblock \href {https://doi.org/10.1561/1500000011} {Opinion mining and
  sentiment analysis}.
\newblock \emph{Foundations and Trends in Information Retrieval},
  2(1–2):1–135.

\bibitem[{Paszke et~al.(2017)Paszke, Gross, Chintala, Chanan, Yang, DeVito,
  Lin, Desmaison, Antiga, and Lerer}]{paszke2017automatic}
Adam Paszke, Sam Gross, Soumith Chintala, Gregory Chanan, Edward Yang, Zachary
  DeVito, Zeming Lin, Alban Desmaison, Luca Antiga, and Adam Lerer. 2017.
\newblock \href
  {https://www.bibsonomy.org/bibtex/2d9d4911f0310e65b1d54ff4c13f11aad/ross_mck}
  {Automatic {D}ifferentiation in {P}y{T}orch}.
\newblock In \emph{NIPS 2017 Workshop on Autodiff}.

\bibitem[{Philips(1985)}]{philips1985william}
Susan~U Philips. 1985.
\newblock William {M.} {O}'{B}arr, {L}inguistic {E}vidence: {L}anguage, power,
  and strategy in the courtroom ({S}tudies on {L}aw and {S}ocial {C}ontrol).
\newblock \emph{Language in Society}, 14(1):113--117.

\bibitem[{Pomerleau and Rao(2017)}]{pomerleau2017fake}
Dean Pomerleau and Delip Rao. 2017.
\newblock \href {http://www.fakenewschallenge.org/} {The {F}ake {N}ews
  {C}hallenge: Exploring how {A}rtificial {I}ntelligence technologies could be
  leveraged to combat {F}ake {N}ews}.
\newblock \emph{Fake News Challenge}.

\bibitem[{Pouran Ben~Veyseh et~al.(2019)Pouran Ben~Veyseh, Nguyen, and
  Dou}]{pouran-ben-veyseh-etal-2019-graph}
Amir Pouran Ben~Veyseh, Thien~Huu Nguyen, and Dejing Dou. 2019.
\newblock \href {https://doi.org/10.18653/v1/P19-1432} {Graph based neural
  networks for event factuality prediction using syntactic and semantic
  structures}.
\newblock In \emph{Proceedings of the 57th Annual Meeting of the Association
  for Computational Linguistics}, pages 4393--4399, Florence, Italy.
  Association for Computational Linguistics.

\bibitem[{Prabhakaran et~al.(2015)Prabhakaran, By, Hirschberg, Rambow, Shaikh,
  Strzalkowski, Tracey, Arrigo, Basu, Clark, Dalton, Diab, Guthrie, Prokofieva,
  Strassel, Werner, Wilks, and Wiebe}]{prabhakaran2015new}
Vinodkumar Prabhakaran, Tomas By, Julia Hirschberg, Owen Rambow, Samira Shaikh,
  Tomek Strzalkowski, Jennifer Tracey, Michael Arrigo, Rupayan Basu, Micah
  Clark, Adam Dalton, Mona Diab, Louise Guthrie, Anna Prokofieva, Stephanie
  Strassel, Gregory Werner, Yorick Wilks, and Janyce Wiebe. 2015.
\newblock \href {https://doi.org/10.18653/v1/S15-1009} {A new dataset and
  evaluation for {B}elief/{F}actuality}.
\newblock In \emph{Proceedings of the Fourth Joint Conference on Lexical and
  Computational Semantics}, pages 82--91, Denver, Colorado. Association for
  Computational Linguistics.

\bibitem[{Prabhakaran et~al.(2010)Prabhakaran, Rambow, and
  Diab}]{prabhakaran2010automatic}
Vinodkumar Prabhakaran, Owen Rambow, and Mona~T. Diab. 2010.
\newblock \href {https://aclanthology.org/C10-2117} {Automatic {C}ommitted
  {B}elief {T}agging}.
\newblock In \emph{International Conference on Computational Linguistics},
  pages 1014--1022, Beijing, China.

\bibitem[{Qi et~al.(2020)Qi, Zhang, Zhang, Bolton, and Manning}]{qi2020stanza}
Peng Qi, Yuhao Zhang, Yuhui Zhang, Jason Bolton, and Christopher~D Manning.
  2020.
\newblock \href {https://aclanthology.org/2020.acl-demos.14.pdf} {Stanza: A
  {P}ython {N}atural {L}anguage {P}rocessing toolkit for many human languages}.
\newblock \emph{arXiv preprint arXiv:2003.07082}.

\bibitem[{Qian et~al.(2018)Qian, Li, Zhang, Zhou, and Zhu}]{qian-2018-gan}
Zhong Qian, Peifeng Li, Yue Zhang, Guodong Zhou, and Qiaoming Zhu. 2018.
\newblock \href {https://doi.org/10.24963/ijcai.2018/597} {Event {F}actuality
  identification via {G}enerative {A}dversarial {N}etworks with auxiliary
  classification}.
\newblock In \emph{Proceedings of the Twenty-Seventh International Joint
  Conference on Artificial Intelligence, {IJCAI-18}}, pages 4293--4300.
  International Joint Conferences on Artificial Intelligence Organization.

\bibitem[{Qian et~al.(2015)Qian, Li, and Zhu}]{qian-2015-ml}
Zhong Qian, Peifeng Li, and Qiaoming Zhu. 2015.
\newblock \href {https://doi.org/10.1109/IALP.2015.7451542} {A two-step
  approach for {E}vent {F}actuality identification}.
\newblock In \emph{2015 International Conference on Asian Language Processing
  (IALP)}, pages 103--106.

\bibitem[{Raffel et~al.(2020)Raffel, Shazeer, Roberts, Lee, Narang, Matena,
  Zhou, Li, and Liu}]{2020t5}
Colin Raffel, Noam Shazeer, Adam Roberts, Katherine Lee, Sharan Narang, Michael
  Matena, Yanqi Zhou, Wei Li, and Peter~J. Liu. 2020.
\newblock \href {http://jmlr.org/papers/v21/20-074.html} {Exploring the limits
  of transfer learning with a unified text-to-text transformer}.
\newblock \emph{Journal of Machine Learning Research}, 21(140):1--67.

\bibitem[{Rambow et~al.(2016)Rambow, Bauer, Radeva, Alagesan, Katsios,
  Strzalkowski, Cardie, Diab, Arrigo, Tracey, Dalton, and
  Dubbin}]{Rambow2016BeSt}
Owen Rambow, Daniel Bauer, Axinia Radeva, Meenakshi Alagesan, Gregorios~A.
  Katsios, Tomek Strzalkowski, Claire Cardie, Mona~T. Diab, Michael Arrigo,
  Jennifer Tracey, Adam Dalton, and Greg Dubbin. 2016.
\newblock \href
  {https://tac.nist.gov/publications/2016/additional.papers/TAC2016.KBP\_Belief\_and\_Sentiment\_overview.proceedings.pdf}
  {The 2016 {TAC KBP BeSt} evaluation}.
\newblock In \emph{Text Analysis Conference}. {NIST}.

\bibitem[{Rashkin et~al.(2017)Rashkin, Choi, Jang, Volkova, and
  Choi}]{rashkin-etal-2017-truth}
Hannah Rashkin, Eunsol Choi, Jin~Yea Jang, Svitlana Volkova, and Yejin Choi.
  2017.
\newblock \href {https://doi.org/10.18653/v1/D17-1317} {Truth of varying
  shades: Analyzing language in fake news and political fact-checking}.
\newblock In \emph{Proceedings of the 2017 Conference on Empirical Methods in
  Natural Language Processing}, pages 2931--2937, Copenhagen, Denmark.
  Association for Computational Linguistics.

\bibitem[{Reich(2005)}]{reich2005reason}
Robert~B Reich. 2005.
\newblock \emph{Reason: Why liberals will win the battle for America}.
\newblock Vintage.

\bibitem[{Rogers et~al.(2020)Rogers, Kovaleva, and
  Rumshisky}]{rogers-etal-2020-primer}
Anna Rogers, Olga Kovaleva, and Anna Rumshisky. 2020.
\newblock \href {https://doi.org/10.1162/tacl_a_00349} {A primer in
  {BERT}ology: What we know about how {BERT} works}.
\newblock \emph{Transactions of the Association for Computational Linguistics},
  8:842--866.

\bibitem[{Rosenthal(2014)}]{rosenthal2014detecting}
Sara Rosenthal. 2014.
\newblock Detecting influencers in social media discussions.
\newblock \emph{XRDS: Crossroads, The ACM Magazine for Students}, 21(1):40--45.

\bibitem[{Rudinger et~al.(2018{\natexlab{a}})Rudinger, Teichert, Culkin, Zhang,
  and Van~Durme}]{rudinger-etal-2018-neural-davidsonian}
Rachel Rudinger, Adam Teichert, Ryan Culkin, Sheng Zhang, and Benjamin
  Van~Durme. 2018{\natexlab{a}}.
\newblock \href {https://doi.org/10.18653/v1/D18-1114} {Neural-{D}avidsonian
  {S}emantic {P}roto-{R}ole labeling}.
\newblock In \emph{Proceedings of the 2018 Conference on Empirical Methods in
  Natural Language Processing}, pages 944--955, Brussels, Belgium. Association
  for Computational Linguistics.

\bibitem[{Rudinger et~al.(2018{\natexlab{b}})Rudinger, White, and
  Van~Durme}]{rudinger2018neural}
Rachel Rudinger, Aaron~Steven White, and Benjamin Van~Durme.
  2018{\natexlab{b}}.
\newblock \href {https://doi.org/10.18653/v1/N18-1067} {Neural models of
  {F}actuality}.
\newblock In \emph{Proceedings of the 2018 Conference of the North {A}merican
  Chapter of the Association for Computational Linguistics: Human Language
  Technologies, Volume 1 (Long Papers)}, pages 731--744, New Orleans,
  Louisiana. Association for Computational Linguistics.

\bibitem[{Saur{\'\i} and Pustejovsky(2009)}]{sauri2009factbank}
Roser Saur{\'\i} and James Pustejovsky. 2009.
\newblock Fact{B}ank: {A} corpus annotated with {E}vent {F}actuality.
\newblock \emph{Language Resources and Evaluation}, 43(3).

\bibitem[{Saur{\'\i} and Pustejovsky(2012)}]{sauri2012you}
Roser Saur{\'\i} and James Pustejovsky. 2012.
\newblock \href {https://doi.org/10.1162/COLI_a_00096} {Are you sure that this
  happened? {A}ssessing the factuality degree of events in text}.
\newblock \emph{Computational Linguistics}, 38(2):261--299.

\bibitem[{Shaffer(1981)}]{shaffer1981balance}
Stephen~D Shaffer. 1981.
\newblock Balance {T}heory and {P}olitical {C}ognitions.
\newblock \emph{American Politics Quarterly}, 9(3):291--320.

\bibitem[{Shapiro(2019)}]{shapiro2019facts}
Ben Shapiro. 2019.
\newblock \emph{Facts don’t care about your feelings}.
\newblock Creators Publishing.

\bibitem[{Shi et~al.(2016)Shi, Padhi, and Knight}]{shi-etal-2016-string}
Xing Shi, Inkit Padhi, and Kevin Knight. 2016.
\newblock \href {https://doi.org/10.18653/v1/D16-1159} {Does string-based
  neural {MT} learn source syntax?}
\newblock In \emph{Proceedings of the 2016 Conference on Empirical Methods in
  Natural Language Processing}, pages 1526--1534, Austin, Texas. Association
  for Computational Linguistics.

\bibitem[{Sim et~al.(2013)Sim, Acree, Gross, and Smith}]{sim2013measuring}
Yanchuan Sim, Brice D.~L. Acree, Justin~H. Gross, and Noah~A. Smith. 2013.
\newblock \href {https://aclanthology.org/D13-1010} {Measuring ideological
  proportions in political speeches}.
\newblock In \emph{Proceedings of the 2013 Conference on Empirical Methods in
  Natural Language Processing}, pages 91--101, Seattle, Washington, USA.
  Association for Computational Linguistics.

\bibitem[{Soni et~al.(2014)Soni, Mitra, Gilbert, and
  Eisenstein}]{soni2014modeling}
Sandeep Soni, Tanushree Mitra, Eric Gilbert, and Jacob Eisenstein. 2014.
\newblock \href {https://doi.org/10.3115/v1/P14-2068} {Modeling factuality
  judgments in social media text}.
\newblock In \emph{Proceedings of the 52nd Annual Meeting of the Association
  for Computational Linguistics (Volume 2: Short Papers)}, pages 415--420,
  Baltimore, Maryland. Association for Computational Linguistics.

\bibitem[{Stanovsky et~al.(2017)Stanovsky, Eckle-Kohler, Puzikov, Dagan, and
  Gurevych}]{stanovsky2017integrating}
Gabriel Stanovsky, Judith Eckle-Kohler, Yevgeniy Puzikov, Ido Dagan, and Iryna
  Gurevych. 2017.
\newblock \href {https://doi.org/10.18653/v1/P17-2056} {Integrating deep
  linguistic features in factuality prediction over unified datasets}.
\newblock In \emph{Proceedings of the 55th Annual Meeting of the Association
  for Computational Linguistics (Volume 2: Short Papers)}, pages 352--357,
  Vancouver, Canada. Association for Computational Linguistics.

\bibitem[{Swamy et~al.(2017)Swamy, Ritter, and de~Marneffe}]{swamy2017have}
Sandesh Swamy, Alan Ritter, and Marie-Catherine de~Marneffe. 2017.
\newblock \href {https://doi.org/10.18653/v1/D17-1166} {{``}i have a feeling
  trump will win..................{''}: Forecasting winners and losers from
  user predictions on {T}witter}.
\newblock In \emph{Proceedings of the 2017 Conference on Empirical Methods in
  Natural Language Processing}, pages 1583--1592, Copenhagen, Denmark.
  Association for Computational Linguistics.

\bibitem[{Swayamdipta and Rambow(2012)}]{swayamdipta2012pursuit}
Swabha Swayamdipta and Owen Rambow. 2012.
\newblock The pursuit of power and its manifestation in written dialog.
\newblock In \emph{2012 IEEE Sixth International Conference on Semantic
  Computing}, pages 22--29. IEEE.

\bibitem[{TAC-KBP(2016)}]{tackbp-2016}
TAC-KBP. 2016.
\newblock Task description for source/target belief and sentiment evaluation
  ({B}e{S}t) at {TAC} 2016.
\newblock \url{http:
  //www.cs.columbia.edu/~rambow/best-eval-2016/task-spec-v2.4.pdf}.

\bibitem[{Tenney et~al.(2019{\natexlab{a}})Tenney, Das, and
  Pavlick}]{tenney-etal-2019-bert}
Ian Tenney, Dipanjan Das, and Ellie Pavlick. 2019{\natexlab{a}}.
\newblock \href {https://doi.org/10.18653/v1/P19-1452} {{BERT} rediscovers the
  classical {NLP} pipeline}.
\newblock In \emph{Proceedings of the 57th Annual Meeting of the Association
  for Computational Linguistics}, pages 4593--4601, Florence, Italy.
  Association for Computational Linguistics.

\bibitem[{Tenney et~al.(2019{\natexlab{b}})Tenney, Xia, Chen, Wang, Poliak,
  McCoy, Kim, Durme, Bowman, Das, and Pavlick}]{tenney2018what}
Ian Tenney, Patrick Xia, Berlin Chen, Alex Wang, Adam Poliak, R~Thomas McCoy,
  Najoung Kim, Benjamin~Van Durme, Sam Bowman, Dipanjan Das, and Ellie Pavlick.
  2019{\natexlab{b}}.
\newblock \href {https://openreview.net/forum?id=SJzSgnRcKX} {What do you learn
  from context? {P}robing for sentence structure in contextualized word
  representations}.
\newblock In \emph{International Conference on Learning Representations}.

\bibitem[{Thorne et~al.(2018)Thorne, Vlachos, Christodoulopoulos, and
  Mittal}]{thorne2018fever}
James Thorne, Andreas Vlachos, Christos Christodoulopoulos, and Arpit Mittal.
  2018.
\newblock \href {http://arxiv.org/abs/1803.05355} {{FEVER:} {A} large-scale
  dataset for {F}act {E}xtraction and {VER}ification}.
\newblock \emph{CoRR}, abs/1803.05355.

\bibitem[{Toulmin(1958)}]{toulmin-1958-the}
Stephen~E. Toulmin. 1958.
\newblock \emph{The Uses of Argument}.
\newblock Cambridge University Press.

\bibitem[{Trautmann et~al.(2020)Trautmann, Daxenberger, Stab, Sch{\"{u}}tze,
  and Gurevych}]{DBLP:conf/aaai/TrautmannDSSG20}
Dietrich Trautmann, Johannes Daxenberger, Christian Stab, Hinrich
  Sch{\"{u}}tze, and Iryna Gurevych. 2020.
\newblock \href {https://aaai.org/ojs/index.php/AAAI/article/view/6438}
  {Fine-grained argument unit recognition and classification}.
\newblock In \emph{The Thirty-Fourth {AAAI} Conference on Artificial
  Intelligence, {AAAI} 2020, The Thirty-Second Innovative Applications of
  Artificial Intelligence Conference, {IAAI} 2020, The Tenth {AAAI} Symposium
  on Educational Advances in Artificial Intelligence, {EAAI} 2020, New York,
  NY, USA, February 7-12, 2020}, pages 9048--9056. {AAAI} Press.

\bibitem[{Vigus et~al.(2019)Vigus, Van~Gysel, and
  Croft}]{vigus-2019-dependency}
Meagan Vigus, Jens E.~L. Van~Gysel, and William Croft. 2019.
\newblock \href {https://doi.org/10.18653/v1/W19-3321} {A dependency structure
  annotation for modality}.
\newblock In \emph{Proceedings of the First International Workshop on Designing
  Meaning Representations}, pages 182--198, Florence, Italy. Association for
  Computational Linguistics.

\bibitem[{Vlachos and Riedel(2014)}]{vlachos2014fact}
Andreas Vlachos and Sebastian Riedel. 2014.
\newblock \href {https://doi.org/10.3115/v1/W14-2508} {Fact {C}hecking: {T}ask
  definition and dataset construction}.
\newblock In \emph{Proceedings of the {ACL} 2014 Workshop on Language
  Technologies and Computational Social Science}, pages 18--22, Baltimore, MD,
  USA. Association for Computational Linguistics.

\bibitem[{Wallach(2014)}]{Wallach2014CSSFAT}
Hanna Wallach. 2014.
\newblock \href
  {https://hannawallach.medium.com/big-data-machine-learning-and-the-social-sciences-927a8e20460d}
  {Big data, machine learning, and the social sciences: Fairness,
  accountability, and transparency}.
\newblock Talk at {NIPS 2014 Workshop on Fairness, Accountability, and
  Transparency in Machine Learning}.

\bibitem[{Walton et~al.(2008)Walton, Reed, and
  Macagno}]{walton2008argumentation}
Douglas Walton, Christopher Reed, and Fabrizio Macagno. 2008.
\newblock \emph{Argumentation {S}chemes}.
\newblock Cambridge University Press.

\bibitem[{Walton(1996)}]{walton1996argumentation}
Douglas~N Walton. 1996.
\newblock \emph{Argumentation schemes for presumptive reasoning}.
\newblock Psychology Press.

\bibitem[{Wasserman(2004)}]{wasserman2004all}
Larry Wasserman. 2004.
\newblock \emph{All of {S}tatistics: {A} concise course in statistical
  inference}, volume~26.
\newblock Springer.

\bibitem[{White et~al.(2016)White, Reisinger, Sakaguchi, Vieira, Zhang,
  Rudinger, Rawlins, and Van~Durme}]{white2016universal}
Aaron~Steven White, Drew Reisinger, Keisuke Sakaguchi, Tim Vieira, Sheng Zhang,
  Rachel Rudinger, Kyle Rawlins, and Benjamin Van~Durme. 2016.
\newblock \href {https://doi.org/10.18653/v1/D16-1177} {Universal
  {D}ecompositional {S}emantics on {U}niversal {D}ependencies}.
\newblock In \emph{Proceedings of the 2016 Conference on Empirical Methods in
  Natural Language Processing}, pages 1713--1723, Austin, Texas. Association
  for Computational Linguistics.

\bibitem[{Wiebe et~al.(2005)Wiebe, Wilson, and Cardie}]{wiebe2005annotating}
Janyce Wiebe, Theresa Wilson, and Claire Cardie. 2005.
\newblock \href {https://doi.org/https://doi.org/10.1007/s10579-005-7880-9}
  {Annotating expressions of opinions and emotions in language}.
\newblock \emph{Language Resources and Evaluation}, 39(2):165--210.

\bibitem[{Wiseman et~al.(2017)Wiseman, Shieber, and
  Rush}]{wiseman-etal-2017-challenges}
Sam Wiseman, Stuart Shieber, and Alexander Rush. 2017.
\newblock \href {https://doi.org/10.18653/v1/D17-1239} {Challenges in
  data-to-document generation}.
\newblock In \emph{Proceedings of the 2017 Conference on Empirical Methods in
  Natural Language Processing}, pages 2253--2263, Copenhagen, Denmark.
  Association for Computational Linguistics.

\bibitem[{Yao et~al.(2021)Yao, Qiu, Zhao, Min, and Xue}]{yao-2021-mds}
Jiarui Yao, Haoling Qiu, Jin Zhao, Bonan Min, and Nianwen Xue. 2021.
\newblock \href {https://doi.org/10.18653/v1/2021.acl-long.122} {Factuality
  assessment as modal dependency parsing}.
\newblock In \emph{Proceedings of the 59th Annual Meeting of the Association
  for Computational Linguistics and the 11th International Joint Conference on
  Natural Language Processing, {ACL/IJCNLP} 2021, (Volume 1: Long Papers),
  Virtual Event, August 1-6, 2021}, pages 1540--1550. Association for
  Computational Linguistics.

\bibitem[{Zhang et~al.(2021)Zhang, Wu, Katiyar, Weinberger, and
  Artzi}]{zhang2021revisiting}
Tianyi Zhang, Felix Wu, Arzoo Katiyar, Kilian~Q Weinberger, and Yoav Artzi.
  2021.
\newblock \href {https://openreview.net/forum?id=cO1IH43yUF} {Revisiting
  few-sample {BERT} fine-tuning}.
\newblock In \emph{International Conference on Learning Representations}.

\bibitem[{Zubiaga et~al.(2016)Zubiaga, Kochkina, Liakata, Procter, and
  Lukasik}]{zubiaga-etal-2016-stance}
Arkaitz Zubiaga, Elena Kochkina, Maria Liakata, Rob Procter, and Michal
  Lukasik. 2016.
\newblock \href {https://www.aclweb.org/anthology/C16-1230} {Stance
  classification in rumours as a sequential task exploiting the tree structure
  of social media conversations}.
\newblock In \emph{Proceedings of {COLING} 2016, the 26th International
  Conference on Computational Linguistics: Technical Papers}, pages 2438--2448,
  Osaka, Japan. The COLING 2016 Organizing Committee.

\end{thebibliography}
\bibliographystyle{acl_natbib}

\newpage

\appendix
\paragraph{Appendices}
\section{Experimental Details}
\subsection{Implementation Details}
\label{sec:appendix_implementation_details}

 
All our models are implemented with PyTorch $1.9$, using \texttt{roberta-large} (with 1024-dimensional embeddings) accessed from AllenNLP $2.5.1$\ \cite{paszke2017automatic,gardner-etal-2018-allennlp}. We train the models with the Adam
optimizer~\cite{DBLP:journals/corr/KingmaB14}, using at most 20 epochs, batch size 16, and learning rate $5 \times 10^{-6}$, following \citet{zhang2021revisiting} and \citet{mosbach2021on}'s training guidelines. We use an early stopping rule if the validation loss does not reduce for more than two epochs; this typically ends training in $5-6$ epochs. We report macro-averaged precision, recall, and F1 over the original train-test set splits of \factbank.
Since fine-tuning BERT (and its variants) can be unstable on small datasets~\cite{dodge-jesse-etal-2020-fine}, we report average performance over
five random restarts for each model.
To fine-tune BERT and RoBERTa models, we start with the pre-trained language model, updating both the task-specific layer and all parameters of the language model. 

\subsection{Significance Testing}
\label{sec:appendix_significance_testing}
We use a nonparametric bootstrap~\cite[ch.~8]{wasserman2004all}
to infer confidence intervals for an individual model's performance metric (precision, recall, F1) and hypothesis testing between pairs of models.
We utilize $10^{4}$ bootstrap samples of sentences for source and event identification models and $10^{4}$ bootstrap samples of epistemic stance tuples for stance classifier in \factbank's test set to report $95\%$ bootstrap confidence intervals (CI), via the normal interval method~\cite[ch.~8.3]{wasserman2004all},
and compare models with a bootstrap two-sided hypothesis test
to calculate a $p$-value for the null hypothesis of two models having an equal macro-averaged F1 score~\cite{mackinnon2009bootstrap}.\footnote{MacKinnon's bootstrap hypothesis test has subtle differences from \citet{berg-kirkpatrick-etal-2012-empirical}'s in the NLP litearture; we find MacKinnon's theoretical justification clearer.}

\section{Performance of Source and Event Identification Models} 
\label{sec:appendix_model_performance}

\subsection{Source and Event Identification}
\label{sec:appendix_source_event_model}
Table~\ref{table:source_event_model} mentions performance scores of the source and event identification models. 
\begin{table}[h]
\tiny
\begin{tabular}{lllllll}
\hline
\multicolumn{1}{c}{\multirow{2}{*}{Model}} & \multicolumn{3}{c}{Event} & \multicolumn{3}{c}{Source} \\ \cline{2-4} \cline{5-7} 
\multicolumn{1}{c}{}                       & Prec   & Recall   & F1    & Prec   & Recall   & F1     \\ \hline
CNN (Qian et al., 2018)                    & 86.6   & 82.8     & 84.6  & 80.7   & 77.4     & 78.9   \\
RoBERTa (Joint)                            & 84.4   & 87.6     & 86.0  & 81.4   & 62.7     & 70.8   \\
RoBERTa (Individual)                       & 84.1   & 87.2     & 85.6  & 79.7   & 81.2     & 80.5   \\ \hline
\end{tabular}
\caption{Performance of the source and event identification models. Individual classifiers perform better than a combined classifier.}
\label{table:source_event_model}
\end{table}

\subsection{Error Analysis: Correlation with the events denoted by verb \textit{"say"}}
We conducted an error analysis of our source identification model. We tested the model to examine whether the model understands the notion of source or merely associates the notion of source with presence of vents denoted by verb “say” in a given sentence. Table~\ref{table:source_error_analysis} demonstrates that the model does not merely rely on presence or absence of such events.
\begin{table}[h]
\centering
\scriptsize
\begin{tabular}{lllll}
\hline
\textbf{“Say”} & \textbf{F1} & \textbf{Precision} & \textbf{Recall} & \textbf{\#sentences} \\ \hline
Present        & 84.6        & 86.4               & 82.9            & 147                  \\
Absent         & 65.2        & 58.4               & 73.8            & 269                  \\ \hline
\end{tabular}
\caption{Source Error Analysis}
\label{table:source_error_analysis}
\end{table}

\section{Performance of Epistemic Stance Classifier} 
\label{sec:appendix_stance_classifier}
\subsection{Error Analysis: Negative Polarity Items}
The \emph{CT-} class is the most rare in \factbank, and it is useful to identify for a possible future use case of finding disagreements in text.  For corpus exploration, an alternative to our model could be to simply view sentences with
explicit negative polarity items (NPIs); such sentences\footnote{Using an NPI list of: 
\emph{no}, \emph{not}, \emph{n't},
\emph{never}, \emph{nobody}, \emph{none} } indeed contain a large majority (88.2$\%$) of \factbank's gold standard \emph{CT-} tuples. They are still uncommon within NPI-containing sentences (13.5$\%$ of such tuples are \emph{CT-}),
and quite rare within sentences without NPIs (0.33$\%$ of such tuples are \emph{CT-}).
For this challenging CT- class, the model attains a F1 score of 78.4$\%$. To examine the model performance on CT- class in political domain, we qualitatively analyzed correct classifications. We observe that the model exhibits ability to deal with complex connections between negation-bearing constructions like \emph{Unable to}, \emph{refuse}, etc. (Table \ref{tab:npi_examples}).

\begin{table*}[!htp]\centering
\scriptsize
\begin{tabular}{p{0.97\linewidth}}\toprule
$\bullet$ \textbf{[Author]$_{s}$}:~Unable to \textbf{\underline{reach}$_{e}$} Russo in the era before cell phones, the House Speaker, Jim Wright, kept the vote open for some twenty minutes while an aide coaxed a member to change his vote to yes. \\
$\bullet$ Author:~\textbf{[John Boehner]$_{s}$},~the Speaker of the House, refused to \textbf{\underline{address}$_{e}$} immigration reform in 2013. \\
$\bullet$ Author:~\textbf{[People]$_{s}$} are beginning to move worlds apart and find it increasingly difficult to~\textbf{\underline{establish}$_{e}$}~common ground. \\
$\bullet$ \textbf{[Author]$_{s}$}:~Although still incapable of actually \textbf{\underline{cutting}$_{e}$}~spending, except for needed defense, conservative leaders imply our national crisis is merely some budgeting blunder remediable through a balanced budget amendment to the Constitution. \\
\bottomrule
\end{tabular}
\caption{Examples of \emph{CT-} epistemic stances, in sentences without explicit NPIs in \polibelief, that BERT correctly predicts; sources are highlighted in bold, and events are underlined.}\label{tab:npi_examples}
\end{table*}

\section{External Validity: A Case Study on Hedging and power}
\label{sec:appendix_hedging}
\noindent
\citet{jalilifar2011power} examine the relationship between an author's perceived political power and their expressed commitment to their beliefs.
While hedging and hesitations have been utilized to measure lack of commitment~\cite{philips1985william}, political discourse can feature many more strategies beyond a simple lexicon of hedge words, such as indirect speech acts, hypothetical if-then clauses, or framing claims as questions~\cite{fraser2010hedging}. Thus, analyzing hedging requires understanding of syntactic contexts within which claims are expressed, which our model can tackle. We establish the external validity of our proposed epistemic stance framework by computationally replicating the findings of \citet{jalilifar2011power}'s manual content analysis. To ensure the external validity of our proposed epistemic stance framework, we  computationally replicate the findings of \citet{jalilifar2011power}'s manual content analysis.

The study examines transcripts of topically similar television interviews of three political figures,
George W.\ Bush (at the time, incumbent U.S.\ president), 
Jimmy Carter (former U.S.\ president),
 and David Coltart (founding member of Zimbabwe’s main opposition party).\footnote{Authors also analyzed interviews by U.S.\ politician Sarah Palin, but we these transcripts were not available at the provided URL.} 
For each interview transcript, we employ our epistemic stance classifier to predict the stance of the political figure (author source) towards all extracted events,
and 
calculate each author's uncertainty level as the fraction of events with a \emph{PR+} or \emph{PS+} epistemic stance.


We find the same ordering of commitment as the previous work:
Bush using the fewest uncertain \textit{PR+/PS+} stances ($5.41\%$),
with progressively more for Carter ($8.32\%$) and Coltart ($12.2\%$).
This follows \citeauthor{jalilifar2011power}'s interpretation of commitment being correlated to power (Bush being the highest status, for example).

\section{Case Study: Belief Holder Identification}
\label{sec:appendix_bh}
\subsection{Details of MMM Corpus}
\label{sec:appendix_mmm}
The MMM, maintained by one of the authors (\emph{anon.\ for review}),
is an example of a researcher-curated ``artisanal data" \cite{Wallach2014CSSFAT} collection, 
common in political science and communication research. Books were chosen according to a number of selection criteria and not as a representative sample of any presumed population of publications. Nominees for consideration include books appearing on best-seller lists from a number of politically-oriented Amazon book categories, mostly under the heading ``Politics \& Government---Ideologies \& Doctrines.''
Additionally, all presidential primary candidates authoring a book during this period were considered, as were other officials (e.g. governors, sheriffs, senators) and ideologues attaining public prominence. Over the course of several years, scholars of American ideology have been invited to nominate additional authors for consideration, as the long-term goal is to maintain as comprehensive as possible a corpus of mass-marketed ideologically-oriented manuscripts. Among nominees, books that were more memoir than manifesto were eliminated, as were books too narrowly focused on a particular policy area.

Books in the MMM were published from 1993 through 2020, with a majority during the Obama presidential administration (233 in 2009-2016),
as well as 57 from the George W. Bush presidency (2001-2008)
and 80 during the Trump presidency (2017-2020).

\subsection{Comparison with NER: Qualitative Examples}
\label{sec:appendix_belief_holders}
Table~\ref{tab:belief holders} describes whether the book's belief holders are recognized as named entities---three of ten are not.

\begin{table}[h]
\centering
\scriptsize
\begin{tabular}{lc|lc}
\hline
\textbf{\begin{tabular}[l]{@{}l@{}}Belief \\ Holder\end{tabular}} &
  \textbf{\begin{tabular}[l]{@{}l@{}}Detected \\ by NER?\end{tabular}} &
  \textbf{\begin{tabular}[l]{@{}l@{}}Belief \\ Holder\end{tabular}} &
  \textbf{\begin{tabular}[l]{@{}l@{}}Detected \\ by NER?\end{tabular}} \\ \hline
Media   & Yes & Bernie Sanders  & No \\
Democrats      & Yes  & Right & Yes \\
Donald Trump     & Yes & Republicans  & No \\
Left       & No  & Courts      & Yes \\
Conservatives       & Yes  & Joe Biden      & Yes \\ \hline
\end{tabular}
\caption{Top 10 sources detected as belief holders in Ben Shapiro's \emph{Facts Don't Care About Your Feelings}.}
\label{tab:belief holders}
\end{table}

\subsection{Linguistic Analysis of Belief Holders}
\label{sec:appendix_linguistic_analysis}
We identify two interesting linguistic phenomena among belief holders mentions.

\paragraph{Common and Collective Nouns}
Many belief holders can also be described by common nouns, such as a plural form referring to classes of people (or other agents), or collective nouns denoting aggregate entities, including informally defined ones. We show several examples, along with an event toward which they have an epistemic stance.

\begin{enumerate}[label={(\arabic*)}]
    
    
    \item \label{senior} A recent survey of studies published in peer-reviewed scientific journals found that 97 percent of actively publishing climate \textbf{[scientists]$_{s}$} agree that global warming has been \textbf{\underline{caused}$_{e}$} by human activity.~\cite{abdul-2016-writings}

    \item \label{left} The \textbf{[Left]$_{s}$} properly pointed out the widespread problems of racism and sexism in American society in the 1950s — and their diagnosis was to \textbf{\underline{destroy}$_{e}$} the system utterly.~\cite{shapiro2019facts}
    


    \item \label{governemnt} The agents seized rosewood and ebony that the \textbf{[government]$_{s}$} believed was illegally \textbf{\underline{imported}$_{e}$}.~\cite{forbes-2012-freedom}
    
    \item \label{media} The \textbf{[media]$_{s}$} simply asserted that Clinton was \textbf{\underline{beloved}$_{e}$} across the land — despite never being able to get 50 percent of the country to vote for him, even before the country knew about Monica Lewinsky.~\cite{coulter2009guilty}
    
    
    \item Maybe American \textbf{[society]$_{s}$} concluded, at some deep level of collective unconsciousness, that it had to \textbf{\underline{reject}$_{e}$} the previous generation ’s model of strict fathering in favor of nurturing mothering.~\cite{reich2005reason}
    
\end{enumerate}

\vspace{-0.2cm}
\paragraph{Word Sense Disambiguation}
If an entity is described as a belief holder, that can help disambiguate its word sense or entity type.
Our model distinguishes agentive versus non-agentive versions of a geographical locations.  In the following two examples, the locations or ideas ``Europe'' and ``Silicon Valley'' 
are belief holders with opinions toward various future scenarios 
(all with uncommitted \emph{Uu} stances, which FactBank uses for all conditionals and hypotheticals). 
These location entities are treated as agents with political desires and intentions, perhaps more like an organizational or geopolitical NER type, despite the fact that these instances do not represent formally defined or even universally agreed-upon entities.


\begin{enumerate}[resume, label={(\arabic*)}]
    
    
    
    
    \item \label{europe_1} 
    \textbf{[Europe]$_{s}$} sees it [NATO expansion] as a scheme for permanent U.S. hegemony and has decided that if the Americans want to play Romans, let Americans \textbf{\underline{pay}$_{e}$} the costs and \textbf{\underline{take}$_{e}$} the risks.~\cite{pat1999republic} 
    

    
    
    
    \item \label{silicon} "Currently \textbf{[Silicon Valley]$_{s}$} is in the midst of a love affair with BMI, arguing that when robots \textbf{\underline{come}$_{e}$} to \textbf{\underline{take}$_{e}$} all of our jobs, we’re going to \textbf{\underline{need}$_{e}$} stronger redistributive policies to \textbf{\underline{help}$_{e}$} \textbf{\underline{keep}$_{e}$} families afloat," Annie Lowrey, who has a book on the subject coming July 10, wrote in New York magazine.~\cite{beck2018addicted} 
    
    

\end{enumerate}

\noindent
By contrast, ``Europe'' and ``Iowa'' below have no epistemic stances (all edges toward sentence events are \emph{NE}), and the entities are used simply to describe geographic locations.

\begin{enumerate}[resume, label={(\arabic*)}]
    \item \label{europe_2} Napoleon was the dictator of a French state so anticlerical that many in \textbf{[Europe]$_{s}$} speculated that he was the Antichrist.~\cite{dreher2018benedict} 
    
    
    \item \label{iowa} 
    While reporters waited outside in the \textbf{[Iowa]$_{s}$} cold amid a mix-up at one of Trump’s rallies [...]
    \cite{abdul-2016-writings} 
    
\end{enumerate}

\end{document}